%% file: main.tex
\documentclass[sigconf]{acmart}
\usepackage{xspace}
\usepackage{cleveref}
\usepackage{bbm}
\usepackage{multirow}
\usepackage{multicol}
\usepackage{microtype}
\usepackage{booktabs}
\usepackage{adjustbox}
\usepackage[utf8]{inputenc}
\usepackage[ruled,vlined]{algorithm2e}
\usepackage{enumitem}
\usepackage{subfigure}
\usepackage{colortbl}
\usepackage{pifont}%
\usepackage{bbding}
\usepackage{balance}

\definecolor{maroon}{cmyk}{0,0.1,0.01,0.01}
\definecolor{blue}{cmyk}{0.95,0.0,0.2,0.2}
\definecolor{yellow}{cmyk}{0.01,0.0,0.2,0.01}
\definecolor{lightblue}{cmyk}{0.1,0.0,0.02,0.02}

\newcommand{\blue}[1]{\textcolor{blue}{\small #1}}

\crefname{section}{§}{§§}
\Crefname{section}{§}{§§}
\AtBeginDocument{%
  }

\copyrightyear{2023}
\acmYear{2023}
\setcopyright{rightsretained}
\acmConference[KDD '23]{Proceedings of the 29th ACM SIGKDD Conference on Knowledge Discovery and Data Mining}{August 6--10, 2023}{Long Beach, CA, USA}
\acmBooktitle{Proceedings of the 29th ACM SIGKDD Conference on Knowledge Discovery and Data Mining (KDD '23), August 6--10, 2023, Long Beach, CA, USA}
\acmDOI{10.1145/3580305.3599318}
\acmISBN{979-8-4007-0103-0/23/08}

\begin{document}


\title{\ours: Learning from Noisy Labels via Dynamics-Enhanced Generative Modeling}

\author{Yuchen Zhuang}
\email{yczhuang@gatech.edu}
\affiliation{%
  \institution{Georgia Institute of Technology}
  \city{Atlanta}
  \state{GA}
  \country{USA}
  \postcode{30332}
}
\author{Yue Yu}
\email{yueyu@gatech.edu}
\affiliation{%
  \institution{Georgia Institute of Technology}
  \city{Atlanta}
  \state{GA}
  \country{USA}
}

\author{Lingkai Kong}
\email{lkkong@gatech.edu}
\affiliation{%
  \institution{Georgia Institute of Technology}
  \city{Atlanta}
  \state{GA}
  \country{USA}
}

\author{Xiang Chen}
\email{xiangche@adobe.com}
\affiliation{%
  \institution{Adobe Research}
  \city{San Jose}
  \state{CA}
  \country{USA}
}
\author{Chao Zhang}
\email{chaozhang@gatech.edu}
\affiliation{%
  \institution{Georgia Institute of Technology}
  \city{Atlanta}
  \state{GA}
  \country{USA}
}

\renewcommand{\shortauthors}{Yuchen Zhuang, Yue Yu, Lingkai Kong, Xiang Chen, \& Chao Zhang}
\newcommand{\std}[1]{\tiny{$\pm$#1}}
\newcommand{\ours}{\textsc{DyGen}\xspace}
\newcommand{\ycz}[1]{\textcolor{purple}{\small [Yuchen: #1]}}
\newcommand{\tofill}[1]{\textcolor{orange}{\small [TO FILL: #1]}}
\newcommand{\yy}[1]{\textcolor{magenta}{\small [Yue: #1]}}
\newcommand{\zc}[1]{\textcolor{magenta}{\small [Chao: #1]}}
\newcommand{\xc}[1]{\textcolor{magenta}{\small [Xiang: #1]}}

\input{sections/0abstract.tex}




\maketitle

\input{sections/1intro.tex}

\input{sections/2related.tex}

\input{sections/3problem.tex}
\input{sections/observation.tex}

\input{sections/4method.tex}

\input{sections/5experiments.tex}
\input{sections/6conclusion.tex}
\begin{acks}
This work was supported in part by NSF (IIS2008334, IIS-2106961, CAREER IIS-2144338),
ONR (MURI N00014-17-1-2656).
\end{acks}

\bibliographystyle{ACM-Reference-Format}
\balance
\bibliography{mybib}

\newpage
\appendix

\input{appendices/a-temp.tex}
\input{appendices/a-formulas.tex}
\input{appendices/a-datasets.tex}
\input{appendices/a-pattern.tex}
\input{appendices/a-rules.tex}
\input{appendices/a-baseline.tex}
\input{appendices/a-implementation.tex}

\input{appendices/a-calibration.tex}
\input{sections/58parameter.tex}

\end{document}

%% file: sections/0abstract.tex
\begin{abstract}

Learning from noisy labels is a challenge that arises in many real-world applications where training data can contain incorrect or corrupted labels.
When fine-tuning language models with noisy labels, models can easily overfit the label noise, leading to decreased performance.
Most existing methods for learning from noisy labels use static input features for denoising, but these methods are limited by the information they can provide on true label distributions and can result in biased or incorrect predictions.
In this work, we propose the Dynamics-Enhanced Generative Model (\ours), which uses dynamic patterns in the embedding space during the fine-tuning process of language models to improve noisy label predictions. \ours uses the variational auto-encoding framework to infer the posterior distributions of true labels from noisy labels and training dynamics.
Additionally, a co-regularization mechanism is used to minimize the impact of potentially noisy labels and priors. \ours demonstrates an average accuracy improvement of 3.10\% on two synthetic noise datasets and 1.48\% on three real-world noise datasets compared to the previous state-of-the-art.
Extensive experiments and analyses show the effectiveness of each component in \ours.
Our code is available for reproducibility on GitHub\footnote{\url{https://github.com/night-chen/DyGen}}.

\end{abstract}

%% file: sections/1intro.tex
\section{Introduction}\label{para:intro}

\begin{figure}[t]
  \centering
  \includegraphics[width=0.9\linewidth]{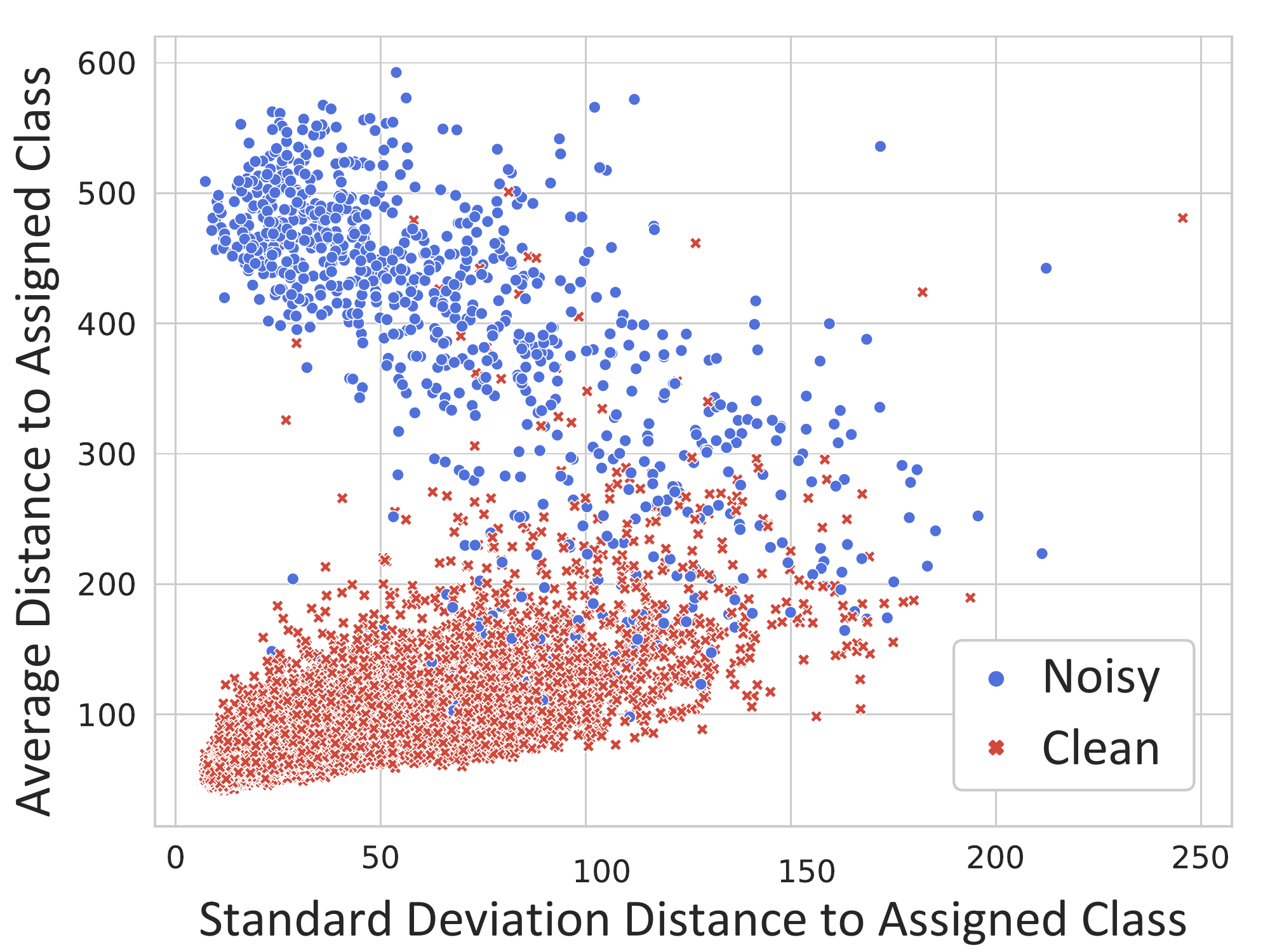}
  \vspace{-1ex}
  \caption{
    The Euclidean distances between instances and their corresponding label cluster centroids in the embedding space on the 20newsgroup dataset with 20\% symmetric noise on labels. The x-axis represents the standard deviation of these distances over epochs during BERT fine-tuning, and the y-axis displays their mean. The patterns of the training dynamics clearly distinguish noisy and clean samples.
  }
  \label{fig:pattern-20news}
  \vspace{-3ex}
\end{figure}

In many applications, collecting clean labeled data can be much more costly compared to obtaining noisy labeled data. Noisy labels can be cheaply obtained in large quantities from sources such as crowdsourcing \cite{winogrande,williams-etal-2018-broad}, web annotations
\cite{blum2003noise,liang2020bond}, labeling rules
\cite{goh2018using,zhang2wrench}, and search engines \cite{Xiao_2015_CVPR, yu2023zero}. Using large-scale noisy labeled data holds the potential of training powerful deep learning models with reduced data curation costs. Particularly, fine-tuning pretrained language models (PLMs) with noisy labels have gained interest for a wide range of text analysis tasks~\cite{swayamdipta-etal-2020-dataset, alt-etal-2020-tacred, zhou-chen-2021-learning}. However, the over-parameterized PLMs, due to their large size, are prone to overfitting the label noise, leading to decreased performance~\cite{pmlr-v70-arpit17a, pmlr-v139-zhang21n, cheng2021learning}. This has become a critical challenge that hinders PLMs from delivering satisfactory results when trained with noisy supervision.

The problem of learning from noisy supervision has been widely studied in the machine learning community. Existing approaches to this issue can be broadly classified into three categories.
1) \emph{Data Cleaning} methods
\cite{pmlr-v97-arazo19a, Li2020DivideMix:, pmlr-v97-yu19b, zhou-chen-2021-learning, pmlr-v119-bahri20a, Nguyen2020SELF:, xia2022sample,xu2023neighborhood}
detect noisy samples using specific criteria such as Area Under Margin ~\cite{NEURIPS2020_c6102b37} and Data Cartography~\cite{swayamdipta-etal-2020-dataset} and remove, reweigh, or correct these samples for subsequent model training.
2) \emph{Regularization} methods design regularized loss functions \cite{xia2021robust, NEURIPS2020_ea89621b, pmlr-v119-han20c, NEURIPS2018_f2925f97, Wang-2019-ICCV, pmlr-v119-ma20c}
or train multiple models to regularize each other
\cite{pmlr-v80-jiang18c, pmlr-v97-yu19b, NEURIPS2018_a19744e2, Tanaka_2018_CVPR, Wei_2020_CVPR, zhou-chen-2021-learning}, with the goal of improving robustness under label noise.
3) \emph{Noise Transition Estimation} methods \cite{Patrini_2017_CVPR, yao2020dual, pmlr-v139-zhang21n, xia2020part, pmlr-v139-berthon21a, NEURIPS2021_23451391}
estimate the transition matrix $p(\tilde{y}|y,\mathbf{x})$ that maps clean labels $y$ to noisy labels $\tilde{y}$, conditioned on input features $\mathbf{x}$.
Noisy Prediction Calibration \cite{pmlr-v162-bae22a}
is a recent approach that models the transition from noisy predictions to the true labels $p(y|\hat{y},\mathbf{x})$.
Each of these categories has its own advantages and drawbacks, and their performance depends on the specific nature of the noise and the input features being used.



A major challenge in existing methods for learning from noisy supervision is their dependence on either the original input features or the embeddings learned with noisy labels. Both scenarios pose limitations when fine-tuning PLMs with noisy labels.
First, the original input features $\mathbf{x}$ from PLMs have limited expressivity and therefore cannot effectively distinguish between noisy and clean labels \cite{li2020sentence}. This can hurt the efficiency of data cleaning methods and the models that learn the noise-to-truth transitions based on $\mathbf{x}$. 
Furthermore, the input features may hold some information about the true labels $y$, however, they only encompass a limited understanding of the relationship between the true labels $y$ and the noisy labels $\tilde{y}$. This limitation leads to a reduced capability for generalization to all types of noise.
Second, fine-tuning PLMs with noisy labels can also hinder the effectiveness of denoising, as label noise can compromise the quality of the learned embeddings. 
Over-fitting to the label noise can cause the model to memorize incorrect labels and mistakenly consider some noisy samples as clean ones, even with metrics in regularization methods during fine-tuning.
This also impedes noise transition estimation methods from accurately modeling the generation of noise.
Consequently, many existing studies are grounded in strong assumptions or are hindered by imprecise noise estimation, resulting in inconsistent performance across varying types of label noise~\cite{song2022learning}.
 
In this work, we have discovered that noisy and clean samples exhibit distinct behaviors in the embedding space during PLM fine-tuning with noisy labels.
During the early stages of fine-tuning, we found that the noisy samples tend to be closer to the cluster associated with the true label $y$. However, as training progresses, these noisy samples are gradually drawn towards the cluster associated with the assigned noisy label $\tilde{y}$.
Therefore, the noisy samples tend to have relatively larger distances to their assigned label clusters due to this training dynamics pattern.
Such dynamic patterns can be quantified by the Euclidean distance between each sample and its assigned cluster center at each training epoch.
In Figure~\ref{fig:pattern-20news}, we visualize the computed distance in the embedding space with the mean (y-axis) and standard deviation (x-axis) over epochs. 
This plot clearly illustrates that noisy samples tend to have larger means and standard deviations as they move from the true label cluster to the noisy label cluster during training.

We thus propose a dynamics-enhanced generative model \ours for denoised fine-tuning of PLMs. 
Our model is based on the observation that noisy and clean samples have different dynamics in the embedding space during the fine-tuning process. 
To take advantage of this dynamic pattern, our model treats the true labels as latent variables and infers them from the dynamic patterns and the noisy labels.
Our model differs from previous generative denoising models \cite{xia2020part, yao2020dual,NEURIPS2021_23451391} in its use of features and modeling of how the features and noisy labels are generated. 
Unlike these previous models, which generate both the noisy label $\hat{y}$ and the input feature $\mathbf{x}$ conditioned on the true label $y$ ($p(\mathbf{x},\tilde{y}|y)$), our model leverages dynamic training patterns $w$ and treats the true label $y$ as the latent encoding of $\hat{y}$. 
This makes it easier to learn, as it only requires generating the noisy label $\hat{y}$, and allows for inference of the posterior $p(y|\hat{y},\mathbf{w})$ using the variational auto-encoding framework. 
Furthermore, we can use the discriminative power of the dynamic patterns to induce the prior distribution $p(y|\mathbf{w})$ of our generative model.
To improve robustness in inferring the true label, we also employ a co-regularization loss that encourages multiple branches of our generative model to reach a consensus for the posterior $p(y|\hat{y},\mathbf{w})$.

We have conducted thorough experiments on two datasets with various synthetic noise types and three datasets from the WRENCH benchmark~\cite{zhang2wrench} with real-world weak label noise. Our method \ours consistently surpasses the state-of-the-art baselines, with an average improvement of $2.13\%$ across various noise types and ratios on both synthetic and real-world datasets. Furthermore, \ours demonstrates remarkable robustness even under extreme label noise ratios, as high as 50\%. Additionally, \ours enhances model calibration by generating predicted probabilities that are more accurately aligned with the true label distribution due to its dynamics-based probabilistic denoising approach. Our contributions are as follows:

\noindent $\bullet$ We have discovered that dynamic training patterns in the hidden embedding space during PLM fine-tuning can effectively distinguish between clean and noisy samples. Utilizing this insight, we have devised a denoised fine-tuning approach for PLMs. To our knowledge, this is the first time that dynamic training patterns are used to achieve robust fine-tuning of PLMs with noisy labels.

\noindent $\bullet$ We design a generative model that models the
reconstruction of the noisy label $\hat{y}$ from the latent true label $y$ and
the training dynamics $\mathbf{w}$. We induce a prior distribution for the latent
variable $y$ based on the dynamics $\mathbf{w}$ and present a training procedure based
on variational auto-encoding.

\noindent $\bullet$ To enhance robustness, we employ multiple branches that co-regularize each other to reach consensus for the posterior $p(y|\hat{y},\mathbf{w})$.

\noindent $\bullet$ We have conducted a comprehensive analysis of the noisy learning problems in text data, covering both synthetic and real-world noise scenarios. Our proposed method, \ours, consistently outperforms other approaches across different types and levels of noise.

%% file: sections/2related.tex
\begin{figure*}[!ht]
	\centering
	\vspace{-2ex}
	\subfigure[20newsgroup]{
		\includegraphics[width=0.48\linewidth]{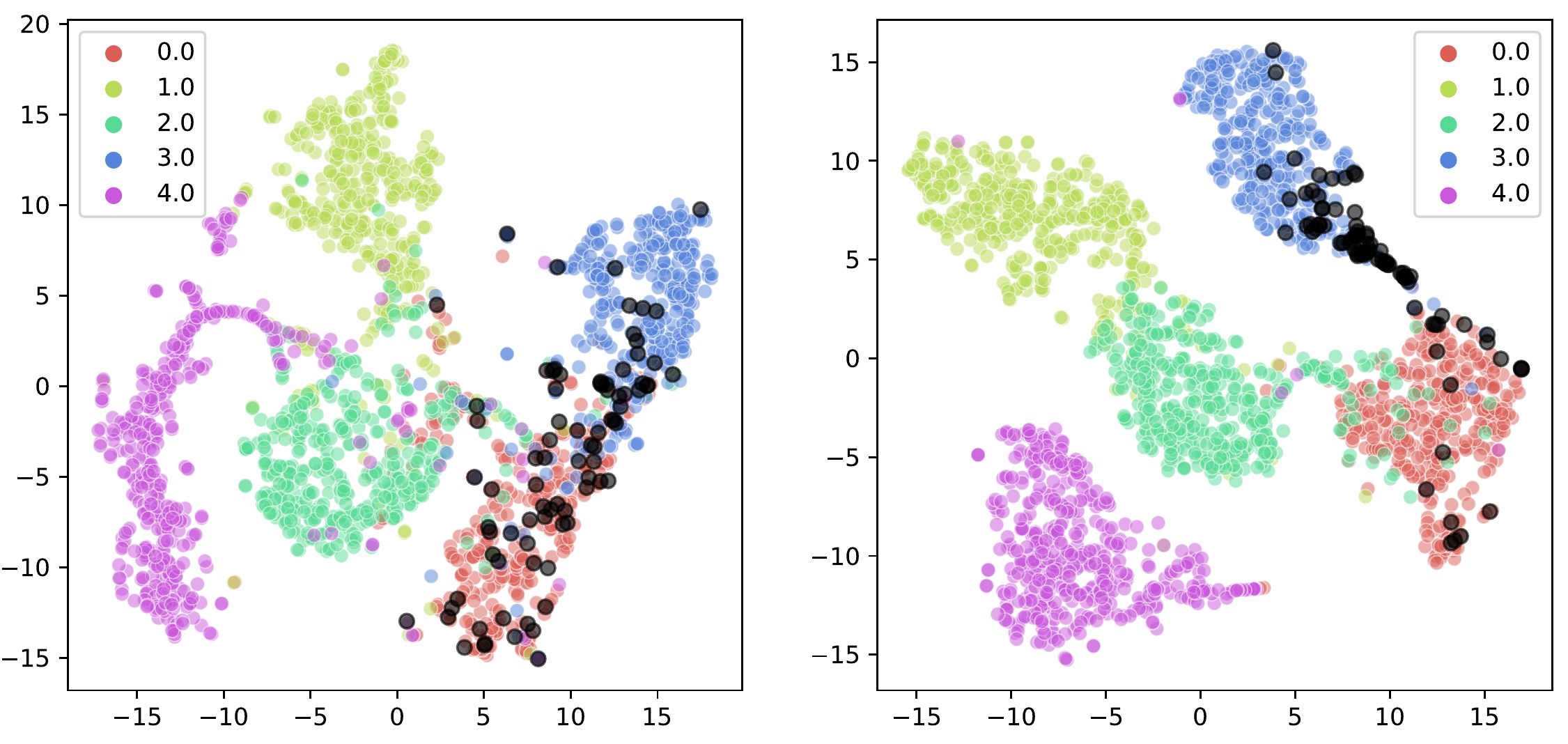}
		\label{fig:ob-20news}
	}  
 \vspace{-2ex}
	\subfigure[ChemProt]{
		\includegraphics[width=0.48\linewidth]{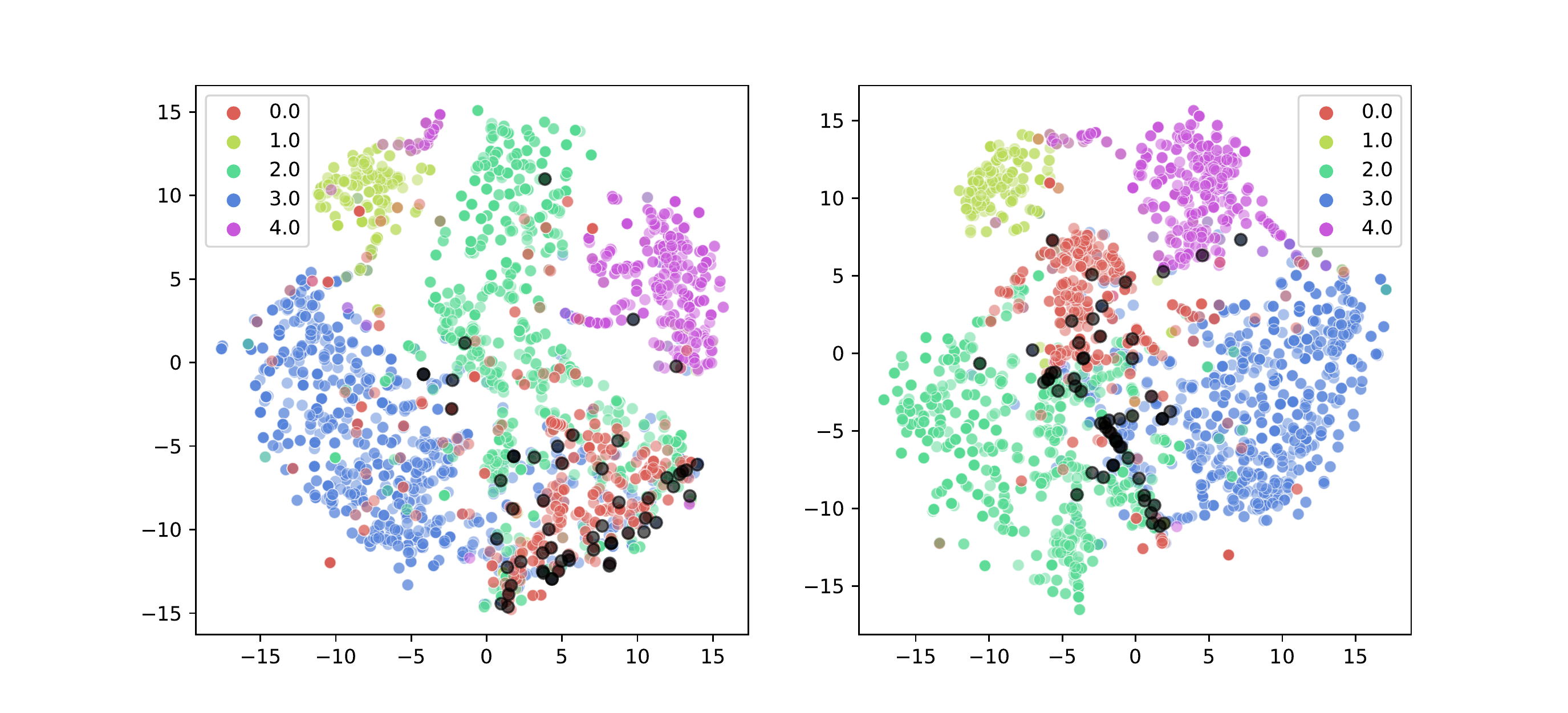}
		\label{fig:ob-chemprot}
	}  
	\vspace{-1ex}
	\caption{Examples of the observed training dynamic pattern in the corrupted dataset, 20newsgroup, and a real-world noisy dataset, ChemProt~\cite{Krallinger2017OverviewOT}. We select 5 categories of the dataset and the black points are the corrupted samples from the true labels (red) to the noisy assigned labels (blue). The left and right figures in each group are t-SNE~\cite{van2008visualizing} visualizations on training embeddings obtained from the $2$-nd and $10$-th epoch, respectively.}\label{fig:intro-example}
	\vspace{-1.5ex}
\end{figure*}

\section{Related Work}
In this section, we briefly introduce several related lines of works on learning from noisy labeled data.

\noindent \textbf{Noise Transition Matrix Estimation.}
Most existing techniques in this line model the generation of noise from true to noisy labels as a transition matrix
\begin{equation}
\begin{aligned}
\mathbf{T}_{j,k}(\mathbf{x})=p(\tilde{y}=j|y=k,\mathbf{x}),\quad \forall j,k=1,\cdots,c.
\end{aligned}
\end{equation}
where $c$ is the total number of classes. If the transition matrix is estimated correctly, classifiers can be trained on noisy data and converge to the optimal solution with theoretical guarantees\cite{li2021provably}. However, the noise transition matrix is difficult to estimate. To improve its modeling, recent works propose various assumptions on the nature of noise.
For example, \cite{Patrini_2017_CVPR, yao2020dual, pmlr-v139-zhang21n} assume the noise is instance-independent, namely $p(\tilde{y}|y,\mathbf{x})=p(\tilde{y}|y)$.
This assumption is often unrealistic for real-world noises, where labeling errors can  depend on the input features $\mathbf{x}$.
\citet{xia2020part} assume that the noise generation is dependent on different parts of an instance; 
\citet{NEURIPS2021_23451391} introduce an auxiliary latent variable $\mathbf{z}$ that works with true label $y$ together to generate the instance feature $\mathbf{x}$.
Nevertheless, these assumptions are too specific and cannot be readily applied to real scenarios where the noise patterns can be diverse and complicated. 
Thus, \citet{pmlr-v162-bae22a} consider the true label $y$ as the latent variable, and infer the posterior:
\begin{equation}
\begin{aligned}
\mathbf{H}_{j,k}(\mathbf{x})=p(y=j|\hat{y}=k,\mathbf{x}),\quad \forall j,k=1,\cdots,c.
\end{aligned}
\label{eq:transmat}
\end{equation}
Our approach adopts the same formulation but improves the generative modeling with training dynamic patterns. These patterns enhance the latent variable modeling and provide more accurate prior and posterior distributions for noisy-to-true label transitions.



\noindent \textbf{Regularization from Multiple Models.}
Deep neural networks are often sensitive to the stochasticity (e.g., weight random initialization and data orders) involved during training. 
This issue is especially exacerbated for noisy label learning, as noisy examples may further disturb model training.  To alleviate this, several works proposed adaptive training strategies that involve \emph{multiple model branches} to improve robustness over noisy labels.
\citet{pmlr-v80-jiang18c} propose two networks, MentorNet and StudentNet, where the MentorNet adopts a reweighting scheme to favor samples that are more likely to be correct to guide the training of the StudentNet.
The Decoupling strategy~\cite{malach2017decoupling} simultaneously trains  two networks and updates parameters only when their predictions disagree.
Co-teaching~\cite{pmlr-v97-yu19b} also involves two networks, where one network learns from the other network's most confident samples.
However, the aforementioned methods only select a part of training examples for training without explicit denoising, and also neglect the information from mislabeled data.
JoCoR~\cite{Wei_2020_CVPR} addresses this problem by incorporating consistency between different models during training, instead of blindly trusting the noisy labels. 
\ours takes advantage of existing approaches by establishing an agreement objective among multiple generative branches with identical structures but different initializations.
By regularizing these branches towards this consensus, we can mitigate the negative effects of noisy labels and potentially imperfect prior knowledge.

\noindent \textbf{Training Dynamics for Data Cleaning.}
Training dynamics depict the behaviors of the model predictions over instances as training progresses.
The main idea of using training dynamics for noisy learning is to consider the patterns as criteria to detect and correct noisy labeled instances.
{Along this line, the most straightforward way is to identify samples with lower training loss as the clean subset~\cite{pmlr-v97-arazo19a, Li2020DivideMix:, pmlr-v97-yu19b,yu2022actune}, but this approach is often too simple and rigid, which end up choosing easy-to-learn examples only.}
To better harness the training dynamics from intermediate iterations, 
\citet{pleiss2020identifying} introduce a new metric that measures the average gap between the logits of a sample's assigned class and its highest non-designated class, where a negative margin suggests the potential label noise;
\citet{swayamdipta-etal-2020-dataset} hypothesize that noisy samples have smaller probabilities in the assigned category during the whole training process.
All these existing metrics for data cleaning are based on heuristics and strong assumptions.
In addition, they are easy to make biased judgments as they depend solely on noisy classifiers' posterior information.

%% file: sections/3problem.tex
\section{Problem Definition}
We study fine-tuning PLMs  with noisy labels for classification, formulated as follows:
given a noisy training dataset $\mathcal{D}_{\rm train}={(\mathbf{x}_i,\tilde{y_i})}_{i=1}^{n}$ consisting of potentially corrupted labels $\tilde{y}$, the ultimate goal is to minimize the true risk $R_L(f):=\mathbb{E}[L(f(x;\theta),y)]$ between the model predictions $f(\mathbf{x};\theta)$ and the underlying true labels $y$, where $L(\cdot)$ is a loss function.
Since the true labels $y$ are not accessible, the only available risk function is the noisy empirical risk $\tilde{R}_L^{emp}(f):=\frac{1}{n}\sum_{i=1}^nL(f(\mathbf{x}_i;\theta),\tilde{y_i})$ based on noisy labels $\tilde{y}$.
Thus, the objective during PLM fine-tuning with noisy labels becomes finding a function that minimizes the true risk $R_L(f)$ through the learning process with the noisy empirical risk $\tilde{R}_L^{emp}(f)$.

%% file: sections/observation.tex
\section{Training Dynamics}\label{sec:dynamics}


\subsection{Training Dynamics in Embedding Space}






We conducted a comprehensive study through over 1,500 experimental trials of fine-tuning various PLMs, including BERT~\cite{devlin-etal-2019-bert}, BioBERT~\cite{lee2020biobert}, PubMedBERT~\cite{pubmedbert}, and RoBERTa~\cite{liu2019roberta}. We used noisy labeled datasets for different NLP benchmarks such as 20newsgroup~\cite{LANG1995331}, AG News~\cite{li-roth-2002-learning, zhang2015character}, ChemProt~\cite{Krallinger2017OverviewOT}, TREC~\cite{Awasthi2020Learning}, and SemEval~\cite{zhou2020nero}, with both synthetic and real-world noise at various ratios.
In our experiments, we modeled the PLM as a two-module model, $f_\theta=g_{\theta_{-1}} \circ h_{\theta_{:-1}}$, where $h_{\theta_{:-1}}$ is the PLM-based encoder and $g_{\theta_{-1}}$ is the final classifier stacked on the encoder.
We optimized the parameters $\theta$ using gradient descent algorithms to minimize the empirical risk $\tilde{R}_L^{emp}(f)$ over $E$ training epochs with noisy labeled data.
During PLM fine-tuning, we observe that the following dynamic patterns consistently occur in the embedding space:

\vspace{1ex}
\textit{When fine-tuning PLMs with noisy labels, noisy samples gradually shift away from the true-label cluster towards their assigned-label cluster in the embedding space, leading to a larger Euclidean distance between the noisy samples and their assigned label cluster centroids across epochs.  Clean samples, on the other hand, exhibit smaller mean and deviation of distances, resulting in a dynamic contrast in their training patterns compared to the noisy samples.}
\vspace{1ex}

Figure~\ref{fig:intro-example} visualizes the dynamic patterns of noisy samples using t-SNE ~\cite{van2008visualizing} on two example settings: 20newsgroup~\cite{LANG1995331} with $20\%$ synthetic symmetric noise and ChemProt~\cite{Krallinger2017OverviewOT} with $22.88\%$ real noise from weak labeling rules.
It shows the embeddings of instances at early and late stages of fine-tuning a BERT-base model.
 Clean samples are represented by colored points, with colors denoting their true labels, and noisy samples are represented by black points, which have moved from their true-label cluster (red) to their assigned-label cluster (blue) during fine-tuning with noisy labels.

The pattern observed is likely due to the \emph{memorization effect} \cite{pmlr-v70-arpit17a,xia2021robust} of deep neural networks, which tend to fit clean label patterns first and then overfit noise. When fine-tuning PLMs with noisy labeled data, all samples tend to remain closer to their true label clusters in the early stages, as PLMs encode semantic knowledge \cite{yu2022cocodr,zhuang-etal-2022-resel} in their embeddings. However, as training continues, the model begins to learn correlations between features and assigned labels, causing noisy samples to gradually move from true-label clusters to assigned-label clusters and overfitting to noise in later epochs. This fitting dynamic still occurs even with large noise ratios, as the randomness of the noise is unlikely to overpower the collective signal of the clean data. Our hypothesis is that this training dynamic will persist as long as there is no systematic bias that dominates the clean data signal.


\subsection{Quantitative Measurements of Pattern}\label{subsec:qp}




To quantify the pattern observed, we measure the Euclidean distances between instance hidden embeddings and the centroids of their assigned label clusters. We fine-tuned the PLM model for $E$ epochs using noisy labeled data and represent the training dynamics of instance $i$ using statistics in the embedding spaces over $E$ epochs. To do this, we first compute the cluster centroids $c^{(e)}_k$ of each class $k$ at each epoch $e$:
\begin{equation}
  \begin{aligned}
    c^{(e)}_k=\frac{1}{n_{\rm train}}\sum_{i=1}^{n_{\rm train}}h_{\theta_{:-1}^{(e)}}(\mathbf{x}_i)\cdot \mathbbm{1}(\tilde{y}_i=k),
  \end{aligned}\label{eq:center}
\end{equation}
where $\theta_{:-1}^{(e)}$ denotes the parameters of the feature encoder $h$ at the $e$-th epoch. Then, we compute the average embedding distance between the samples and the assigned label cluster centroids:
\begin{equation}
  \begin{aligned}
    \mu_i=\frac{1}{E}\sum_{e=1}^E\|h_{\theta_{:-1}^{(e)}}(\mathbf{x}_i),c^{(e)}_{\tilde{y}_i}\|_2.
  \end{aligned}\label{eq:mean}
\end{equation}
In addition, we compute the standard deviation of distances over $E$ epochs to measure the magnitude of distance variations:
\begin{equation}
  \begin{aligned}
    \sigma_i=\sqrt{\frac{1}{E}\sum_{e=1}^E(\|h_{\theta_{:-1}^{(e)}}(\mathbf{x}_i),c^{(e)}_{\tilde{y}_i}\|_2-\mu_i)^2}.
  \end{aligned}\label{eq:std}
\end{equation}
The differences between noisy and clean data are clearly illustrated in Figure~\ref{fig:pattern-20news}. The noisy samples exhibit larger mean and standard deviations in their distances to assigned label clusters compared to clean samples.

%% file: sections/4method.tex
\section{method}
\begin{figure*}[t]
  \centering
  \includegraphics[width=0.9\linewidth]{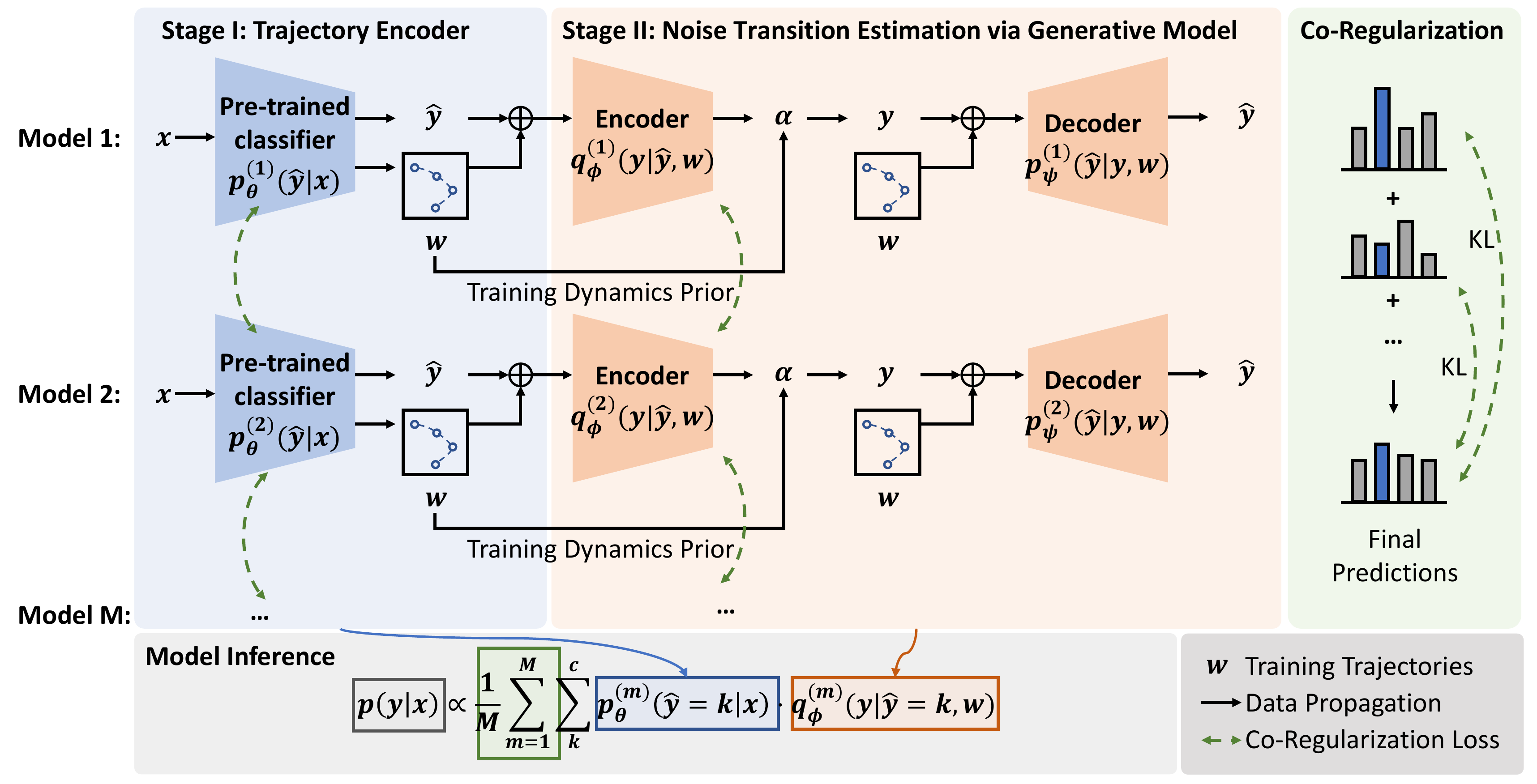}
  \caption{The \ours framework, containing (1) the noisy-supervised model for training trajectory pattern encoding; (2) the generative process, considering true label $y$ as a latent variable and reconstructing $\hat{y}$; (3) co-regularization loss between multiple branches of models; (4) the inference process to predict true labels from noisy predictions. }
  \label{fig:model}
\end{figure*}

For PLM fine-tuning with noisy labels, the ultimate goal is to learn a model that produces the distribution over the true label $y$ for any input $\mathbf{x}$, namely $p(y|\mathbf{x})$.
As we have only noisy labeled data during training, we decompose this objective as:
\begin{equation}
\begin{aligned}
p(y|\mathbf{x})=\sum_{\hat{y}}p(\hat{y}|\mathbf{x})p(y|\hat{y},\mathbf{x}),
\end{aligned}\label{eq:goal}
\end{equation}
where  $\hat{y}$ is the observed noisy label for instance $\mathbf{x}$.

In the above equation, the $p(\hat{y}|\mathbf{x})$ is the biased model learned with the noisy labeled data $\mathcal{D}_{\rm train}$ using standard fine-tuning.
The challenge is to infer the true labels' posterior distribution, $p(y|\hat{y},\mathbf{x})$, which serves as a calibration term that debiases $p(\hat{y}|\mathbf{x})$.
According to the observation in \cref{sec:dynamics}, we propose to use the training dynamics  $\mathbf{w}$
in lieu of $\mathbf{x}$, as $\mathbf{w}$ contains rich information about both noisy predictions $\hat{y}$ and clean labels $y$. 
Based on this insight, the objective is reformulated as:
\begin{equation}
\small
\begin{aligned}
p(y|\mathbf{x})\propto \sum_{\hat{y}}p(\hat{y}|\mathbf{x})p(y|\hat{y},\mathbf{w}).
\end{aligned}\label{eq:newgoal}
\end{equation}


To model the two distributions $p(\hat{y}|\mathbf{x})$ and $p(y|\hat{y},\mathbf{w})$ in Eq.~\ref{eq:newgoal}, we propose a two-stage learning process (also see Figure~\ref{fig:model}): (1) \emph{Stage I}: Learn the standard noisy-supervised model to estimate $p(\hat{y}|\mathbf{x})$ and encode the training trajectories $\mathbf{w}$ as compositions of hidden embeddings obtained from each epoch during fine-tuning;
(2) \emph{Stage II}: Learn the deep generative model to estimate the transition from noisy predictions to true labels and model the function $p(y|\hat{y},\mathbf{w})$. 
The rest of this section describes:
(a) Training trajectory-based deep generative model in \cref{subsec:NPC}, (b) Co-regularization mechanism in \cref{subsec:co-reg}, and (c) Training objective in \cref{sec:obj}.

\input{sections/42generative.tex}

\input{sections/41prior.tex}
\input{sections/43loss.tex}

\input{sections/44objective.tex}

\input{tables/full-alg.tex}

%% file: sections/42generative.tex
\subsection{Deep Generative Model on Training Patterns}\label{subsec:NPC}

\subsubsection{Probabilistic Model Structure}
To model distribution $p(y|\hat{y}, \mathbf{w})$, we introduce a deep generative model that can encode and reconstruct noisy labels $\hat{y}$, conditioned on the training trajectories $\mathbf{w}$.
Specifically, we consider the true labels $y$ as the latent variables and define the following generative process:
first, $y$ is drawn conditioned on the training trajectories $\mathbf{w}$; then, $\hat{y}$ is generated based on $y$ and $\mathbf{w}$.
By following this generative process, we can factorize the joint distribution, modeling the relationship between observations $\hat{y}$ and latent variables $y$:
\begin{equation}
    \begin{aligned}
        p(y,\hat{y}|\mathbf{w})=p(y|\mathbf{w})p(\hat{y}|y,\mathbf{w}).
    \end{aligned}\label{eq:joint}
\end{equation}

Since true labels $y$ are typically assumed to be categorical, we treat $y$ as random probability vectors sampled from a Dirichlet distribution:
\begin{equation}
\begin{aligned}
&y\sim {\rm Dirichlet}(\alpha_\mathbf{w}),\ \hat{y}\sim {\rm Multi}(\pi_{\mathbf{w},y}),
\end{aligned}\label{eq:generative}
\end{equation}
where $\alpha_\mathbf{w}\in\mathbb{R}_+^c$ represents the instance-dependent parameters of the prior probability distribution for all $c$ categories, given a training trajectory $\mathbf{w}$;
${\rm Dirichlet(\alpha_\mathbf{w})}$ is a Dirichlet distribution parameterized by $\alpha_\mathbf{w}$, which is also a conjugate prior of the corresponding multinomial distribution;
and $\pi_{\mathbf{w},y}$ is the probability of selecting a class for the noisy label.

\subsubsection{Dynamics-Based Prior}\label{subsubsec:prior}
Since the prior function $p(y|\mathbf{w})$ in Eq.~\ref{eq:joint} is unknown from the training stage, we approximate it as $p_\theta(y|\mathbf{w})$ using the observed training dynamics patterns (derived in \cref{sec:dynamics}), where $\theta$ is the trajectory encoder parameter in Stage I. 
First, to effectively distinguish between noisy and clean samples, we sum up the mean and standard deviation computed from Eq.\ref{eq:mean} and Eq.\ref{eq:std} as a scoring function: $s_i=\mu_i+\sigma_i$.
Second, we assume that the top $\beta$ percent of instances with the highest $s_i$ are potentially noisy, denoted as $\hat{\mathcal{D}}_{\rm train}^{\rm noisy}$, where $\beta$ is the estimated error rate.
The remaining instances with lower $s_i$ can be considered clean, denoted as $\hat{\mathcal{D}}_{\rm train}^{\rm clean}$.
We use K Nearest Neighbor (KNN) algorithm on $\hat{\mathcal{D}}_{\rm train}^{\rm noisy}$, with $\hat{\mathcal{D}}_{\rm train}^{\rm clean}$ as the reference set to sample $K$ neighbors from.
Third, we combine the most selected labels $\bar{y}$ from neighbors for $\hat{\mathcal{D}}_{\rm train}^{\rm noisy}$ and the remaining assigned labels $\tilde{y}$ for $\hat{\mathcal{D}}_{\rm train}^{\rm clean}$ to update $\mathcal{D}_{\rm train}=\{(\mathbf{x}_i,\tilde{y}_i,y^{\rm prior}_i)\}_{i=1}^{n_{\rm train}}$, where $y_i^{\rm prior}$ indicates the prior knowledge of possible true labels.

With the prior knowledge $y_i^{\rm prior}$, we can define the Dirichlet distribution parameter $\alpha_w$ as:
\begin{equation}
\begin{aligned}
\alpha_w^k=\left\{
\begin{aligned}
&\delta, &k\neq y^{\rm prior}_i\\
&\delta+\rho,&k=y^{\rm prior}_i
\end{aligned}
\right.
,\ k=1,2,\cdots,c.
\end{aligned}\label{eq:prior-para}
\end{equation}
where $\delta$ and $\rho$ are hyper-parameters to setup $\alpha$ for Dirichlet distribution.
The prior function $p_\theta(y|\mathbf{w})$ can then be defined as:
\begin{equation}
    \begin{aligned}
        p_\theta(y|\mathbf{w})={\rm Dirichlet}(\alpha_w).
    \end{aligned}\label{eq:prior}
\end{equation}
Algorithm~\ref{alg:prior} outlines the computation of dynamics-based priors. 

\begin{algorithm}[!ht]
  \begin{small}
	\KwIn{$\mathcal{D}_{\rm train}=\{(\mathbf{x}_i,\tilde{y}_i)\}_{i=1}^{n_{\rm train}}$: noisy training data ; $f_\theta$: noisy supervised model; $E$: Number of $f_\theta$ training epochs.}
	\blue{// Step 0: \textit{Initialization}} \\
    Prepare training trajectory set $\mathcal{S}_{\rm dist}=\emptyset$  \\
	\For{$e = 1, 2, \cdots, E$}{
      {
        \blue{// {Step 1}: \textit{Gather Information for Training Trajectories Encoding.}} \\
        Compute the centroid point of each category on embedding space $c_k^{(e)}$ via Eq.~\ref{eq:center}.\\
        Compute Euclidean Distances between Samples $\mathbf{x}$ and Assigned Class $c_{\hat{y}}^{(e)}$: $\mathcal{S}_{\rm dist}\leftarrow\{\|h_{\theta_{:-1}^{(e)}}(\mathbf{x}),c^{(e)}_{\tilde{y}}\|_2\}$. \\
      }
	}
    \blue{// Step 2: \textit{Compute Scoring Function as Quantitative Pattern.}} \\
    Compute the statistics via Eq.~\ref{eq:mean} and Eq.~\ref{eq:std} on $\mathcal{S}_{\rm dist}$.\\
    Compute the scoring function $s_i=\mu_i+\sigma_i,\ 0\leq i<n_{\rm train}$.\\
    Separate the $\mathcal{D}_{\rm train}$ into possibly-clean set $\hat{\mathcal{D}}_{\rm train}^{\rm clean}$ (smaller $s_i$) and possibly-noisy set $\hat{\mathcal{D}}_{\rm train}^{\rm noisy}$ (larger $s_i$).\\
    \blue{// Step 3: \textit{Generate True Label Prior Information.}} \\
    Apply KNN with $\hat{\mathcal{D}}_{\rm train}^{\rm clean}$ as reference set to correct labels in $\hat{\mathcal{D}}_{\rm train}^{\rm noisy}$ to obtain prior knowledge $\{y_i^{\rm prior}\}_{i=1}^{n_{\rm train}}$.\\
    Compute the parameters to the prior distribution via Eq.~\ref{eq:prior-para}\\
    Compute the approximated prior distribution $p_\theta(y|\mathbf{w})$ via Eq.~\ref{eq:prior}.\\
	\KwOut{The approximated prior distribution $p_\theta(y|\mathbf{w})$.}
  \end{small}
  \caption{Computation of Dynamics-Based Prior. }
  \label{alg:prior}
\end{algorithm}

\subsubsection{Deep Generative Model Architecture}
As our main estimation target $p(y|\hat{y},\mathbf{w})$ in Eq.~\ref{eq:newgoal} is intractable, we apply variational inference to approximate the desired posterior distribution.
To this end, we first introduce a variational distribution $q_\phi(y|\hat{y},\mathbf{x})$. We then minimize the Kullback-Leibler (KL) divergence between the true posterior $p(y|\hat{y},\mathbf{x})$ and the variational distribution $q_\phi(y|\hat{y},\mathbf{x})$.
Specifically, we adopt the structure of Variational Autoencoder (VAE) to parameterize the encoder $q_\phi(y|\hat{y},\mathbf{w})$ and decoder $p_\psi(\hat{y}|y,\mathbf{w})$.
Thus, we can write the KL divergence function:
\begin{equation}
\begin{aligned}
&{\rm KL}(q_\phi(y|\hat{y},\mathbf{w})\|p(y|\hat{y},\mathbf{w}))=\int q_\phi(y|\hat{y},\mathbf{w})\log\frac{q_\phi(y|\hat{y},\mathbf{w})}{p(y|\hat{y},\mathbf{w})}dy\\
&=\log p(\hat{y}|\mathbf{w})-\mathbb{E}_{y\sim q_\phi(y|\hat{y},\mathbf{w})}[\log p_\psi(\hat{y}|y,\mathbf{w})]\\
&\quad +{\rm KL}(q_\phi(y|\hat{y},\mathbf{w})\|p(y|\mathbf{w}))\\
&=\log p(\hat{y}|\mathbf{w})-{\rm ELBO}.
\end{aligned}
\end{equation}
Different from the normal distribution prior in traditional VAE, we apply the reparameterization trick for Dirichlet distribution from Dirichlet VAE~\cite{joo2020dirichlet}:
\begin{equation}\label{eq:elbo}
\begin{aligned}
{\rm ELBO}&=\sum_{k=1}^c (\underbrace{\hat{y}^k\log\hat{y^*}^k+(1-\hat{y}^k)\log(1-\hat{y^*}^k)}_{\rm Reconstruction\ Loss}\\
&\underbrace{-\log\Gamma(\alpha_\mathbf{x}^k)+\log\Gamma(\hat{\alpha}_{\mathbf{x},\hat{y}})-(\hat{\alpha}^k_{\mathbf{x},\hat{y}}-\alpha_\mathbf{x}^k)\psi(\hat{\alpha}^k_{\mathbf{x},\hat{y}})}_{\rm Prior\ Regularizer}),
\end{aligned}
\end{equation}
where $\hat{y^*}$ is the reconstructed $\hat{y}$ after decoder $p_\psi(\hat{y}|y,\mathbf{x})$; $\Gamma(\cdot)$ and $\psi(\cdot)$ represent the gamma and digamma function, respectively. 

\subsubsection{Model Inference}
To perform model inference, we compute $p(y|\mathbf{w},\mathbf{x})$ in Eq.~\ref{eq:newgoal}.
After training the whole architecture with ${\rm ELBO}$ in Eq.~\ref{eq:elbo}, we obtain the posterior distribution from the encoder.
Thus, we can directly use the mode of Dirichlet distribution to compute $H$: 
\begin{equation}
\begin{aligned}
H=q_\phi(y|\hat{y},\mathbf{w})=\frac{\hat{\alpha}_{w,\hat{y}}^y-1}{\sum_{y=1}^c\hat{\alpha}_{w,\hat{y}}^y-c},
\end{aligned}
\label{eq:H}
\end{equation}
where $\hat{\alpha}_{w,\hat{y}}^y$ is the predicted posterior Dirichlet distribution by VAE.
We can then rewrite Eq.~\ref{eq:newgoal} for inference with $q_\phi(y|\hat{y},\mathbf{w})$ in Eq.~\ref{eq:H} :
\begin{equation}
\begin{aligned}
p(y|\mathbf{x})\propto \sum_{k=1}^c q_\phi(y|\hat{y}=k,\mathbf{w})p(\hat{y}=k|\mathbf{x}).
\end{aligned}
\end{equation}

%% file: sections/43loss.tex
\subsection{Co-Regularization Mechanism}\label{subsec:co-reg}

Despite efforts to mitigate the negative impact of noisy samples (\cref{subsec:NPC}), the guidance from labels and prior is still imperfect. Small deviations in $p(\hat{y}|x)$ and $p(y|\hat{y},x)$ could potentially carry over into later stages and affect the overall $p(y|x)$. To address the limitations of imperfect guidance and prevent error propagation, we incorporate multiple branches with identical structures but differing initializations into our model. We use a co-regularization loss across branches to promote consensus and prevent over-reliance on potentially corrupted labels.


The learning process of the co-regularization mechanism involves the learning of $M$ copies of the first-stage model $p_{\theta}^{(m)}(\hat{y}|\mathbf{x})$ and second-stage generative models $q_{\phi}^{(m)}(y|\hat{y},\mathbf{x})$ and $p_{\psi}^{(m)}(\hat{y}|y,\mathbf{x})$, where $m$ ranges from 1 to $M$ ($M>2$).
To begin, we input the instances $\mathbf{x}_i$ into different models to obtain corresponding prediction probabilities $p_i^{(m)}$ of each model $m$. The aggregated probability $q_i$ is then computed by averaging these predictions, $r_i=\frac{1}{M}\sum_{m=1}^Mp_i^{(m)}$, representing the consensus of the models on the true label prediction. The co-regularization loss is calculated as the KL Divergence between the consensus probability $r_i$ and each of the model predicted probabilities $p_i^{(m)}$:

\begin{equation}
  \begin{aligned}
    \ell_{\rm cr}&=\frac{1}{MN}\sum_{i=1}^N\sum_{m=1}^M{\rm KL}(r_i\|p_i^{(m)})\\
                 &=\frac{1}{MN}\sum_{i=1}^N\sum_{m=1}^M\sum_{j=1}^Cr_{ij}\log(\frac{r_{ij}+\epsilon}{p_{ij}^{(m)}+\epsilon}),
  \end{aligned}\label{eq:cr}
\end{equation}
where $\epsilon$ indicates a small positive number to avoid division by zero.
Specifically, for Stage I, we define the consensus probabilities and co-regularization loss as follows:
\begin{equation}
  \begin{aligned}
    r(\hat{y}_i|\mathbf{x}_i)&=\frac{1}{M}\sum_{m=1}^Mp_\theta^{(m)}(\hat{y}_i|\mathbf{x}_i),\\
    \ell_{\rm cr-1}&=\frac{1}{MN}\sum_{i=1}^N\sum_{m=1}^M{\rm KL}(r(\hat{y}_i|\mathbf{x}_i)\|p_\theta^{(m)}(\hat{y}_i|\mathbf{x}_i)).
  \end{aligned}\label{eq:cr1}
\end{equation}
Similarly, the consensus probabilities and co-regularization loss for the deep generative model in Stage II can be represented as:
\begin{equation}
  \begin{aligned}
    r(y|\hat{y}_i,\mathbf{w}_i)&=\frac{1}{M}\sum_{m=1}^Mq_\phi^{(m)}(y_i|\hat{y}_i,\mathbf{w}_i),\\
    \ell_{cr-2}&=\frac{1}{MN}\sum_{i=1}^N\sum_{m=1}^M{\rm KL}(r(y_i|\hat{y}_i,\mathbf{w}_i)\|q_\phi^{(m)}(y_i|\hat{y}_i,\mathbf{w}_i)).
  \end{aligned}\label{eq:cr2}
\end{equation}

%% file: sections/44objective.tex
\subsection{Training Objective}\label{sec:obj}
The training objective of \ours is to optimize a joint loss that combines the task-specific loss and the co-regularization loss. For the first stage of dynamics pattern encoding, the task-specific loss $\ell_{\rm task-1}$ is computed as the cross-entropy loss for classification:
\begin{equation}
\begin{aligned}
\ell_{\rm task-1}=\frac{1}{M}\sum_{m=1}^M\sum_{k=1}^c -\tilde{y}^k\log\hat{y}^{(m)}, 
\end{aligned}\label{eq:task1}
\end{equation}
where $\hat{y}^{(m)}$ indicates the predicted label from the $m$-th model. 
Similarly, task-specific loss $\ell_{\rm task-2}$ for the second stage is calculated as the average negative ELBO in Eq.~\ref{eq:elbo}, across all branches of the model.
Consequently, the training objectives of Stage I and II are defined as:
\begin{equation}
\begin{aligned}
\ell_1&=\ell_{task-1}+\lambda_1\ell_{cr-1},\\
\ell_2&=\ell_{task-2}+\lambda_2\ell_{cr-2},\\
\end{aligned}
\end{equation}
where $\lambda_1$ and $\lambda_2$ are positive hyper-parameters.

To further enhance the training process, we implement a warm-up phase for $\lambda_1$ and $\lambda_2$. 
During this phase, $\lambda$ is temporarily set to $0$ to guarantee proper model initialization. 
Upon completion of the warm-up phase, $\lambda$ will return to its pre-determined positive value. 
Finally, to obtain the final model predictions, the outputs from each model branch are averaged.:
\begin{equation}
\begin{aligned}
p(y|\mathbf{x})\propto\sum_{\hat{y}}\frac{1}{M}p_\theta^{(m)}(\hat{y}|\mathbf{x})q_\phi^{(m)}(y|\hat{y},\mathbf{w}).
\end{aligned}\label{eq:infer}
\end{equation}
We present the learning  procedure of \ours in Algorithm~\ref{alg:fullalg}.

%% file: tables/full-alg.tex
\begin{algorithm}[htb]
	\begin{small}
	\KwIn{$\mathcal{D}_{\rm train}=\{(x_i,\tilde{y}_i)\}_{i=1}^{n_{\rm train}}$: noisy training data ; $\theta$: model parameter for pattern encoder; $E$: number of stage I training epochs; $\phi$ and $\psi$: model parameters in VAE; $T$: number of stage II training iterations; $\gamma$: warm-up ratio; $\lambda_1$ and $\lambda_2$: hyper-parameters; $M$: the number of model branches.}
	\blue{// Step 1: \textit{Encode Training Dynamics Pattern.}} \\ 
	\For{$e = 1, 2, \cdots, E$}{ 
		{
		  \For{$m = 1, 2, \cdots, M$}{
                Compute  $p_\theta^{(m)}(\hat{y}|x)$.\\
            }
            Compute the task-specific loss $\ell_{\rm task-1}$ via Eq.~\ref{eq:task1}.\\
            Compute the co-regularization loss $\ell_{\rm cr-1}$ via Eq.~\ref{eq:cr1}.\\
            \If{$e<\gamma\times E$}{
                $\ell_1=\ell_{\rm task-1}$.
            }\Else{
                $\ell_1=\ell_{\rm task-1}+\lambda_1\ell_{\rm cr-1}$.
            }
		    Update the stage I model parameters $\theta \leftarrow \nabla \ell_1$.
		}
	}
 \blue{// Step 2: \textit{Compute Dynamics-Enhanced Prior.}} \\
\For{$m = 1, 2, \cdots, M$}{
    Compute dynamics-enhanced prior for each model branch $y^{{\rm prior},(m)}$.
}
 \blue{// Step 3: \textit{Generative Model for NPC.}} \\
 \For{$t = 1, 2, \cdots, T$}{
    \For{$m = 1, 2, \cdots, M$}{
        Compute $q_\phi^{(m)}(y|\hat{y},x)$ and $p_\psi^{(m)}(\hat{y}|y,x)$.\\
    }
    Compute the task-specific loss $\ell_{\rm task-2}$ via Eq.~\ref{eq:elbo}.\\
    Compute the co-regularization loss $\ell_{\rm cr-2}$ via Eq.~\ref{eq:cr2}.\\
    \If{$t<\gamma\times T$}{
        $\ell_2=\ell_{\rm task-2}$.
    }\Else{
        $\ell_2=\ell_{\rm task-2}+\lambda_2\ell_{\rm cr-2}$.
    }
    Update the Stage II model parameters $\{\phi,\psi\}\leftarrow \nabla \ell_2$
 }
 \blue{// Step 4: \textit{Model Inference.}} \\
 Compute and average the predictions as $p(y|x,w)$ via Eq.~\ref{eq:infer}.\\
	\KwOut{The inferred true label for each instance .}
	\end{small}
	\caption{Training procedure of \ours. }
	\label{alg:fullalg}
\end{algorithm}

%% file: sections/5experiments.tex
\section{Experiments}

\input{sections/50setup.tex}
\input{tables/tab-syn.tex}
\input{tables/tab-real.tex}

\input{sections/51main.tex}

\begin{figure}[t]
  \centering
  \includegraphics[width=\linewidth]{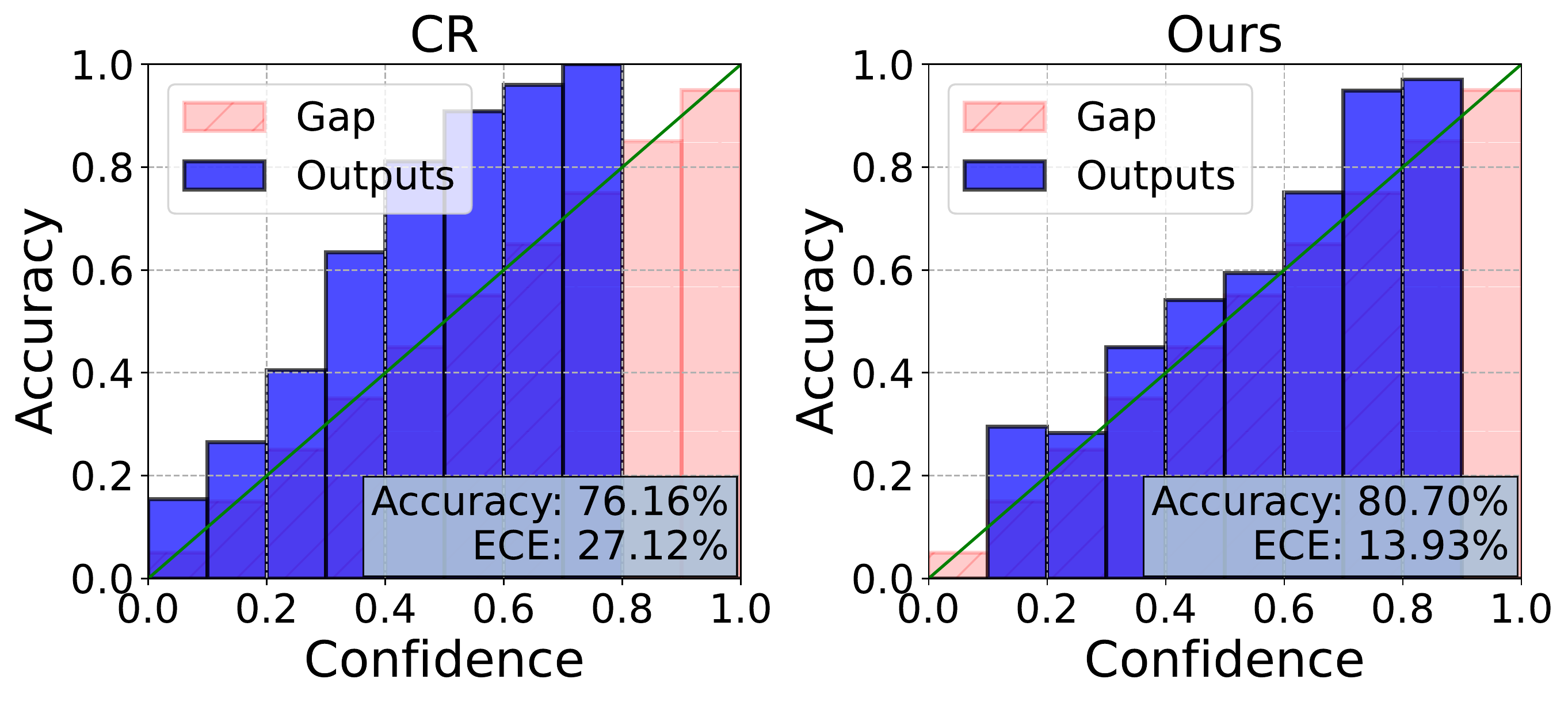}
    \vspace{-2ex}
  \caption{The reliability diagrams of CR baseline and \ours on 20newsgroup with $40\%$ symmetric noise.}
  \vspace{-2ex}
  \label{fig:calibration}
\end{figure}

\begin{figure}[t]
  \centering
  \includegraphics[width=0.9\linewidth]{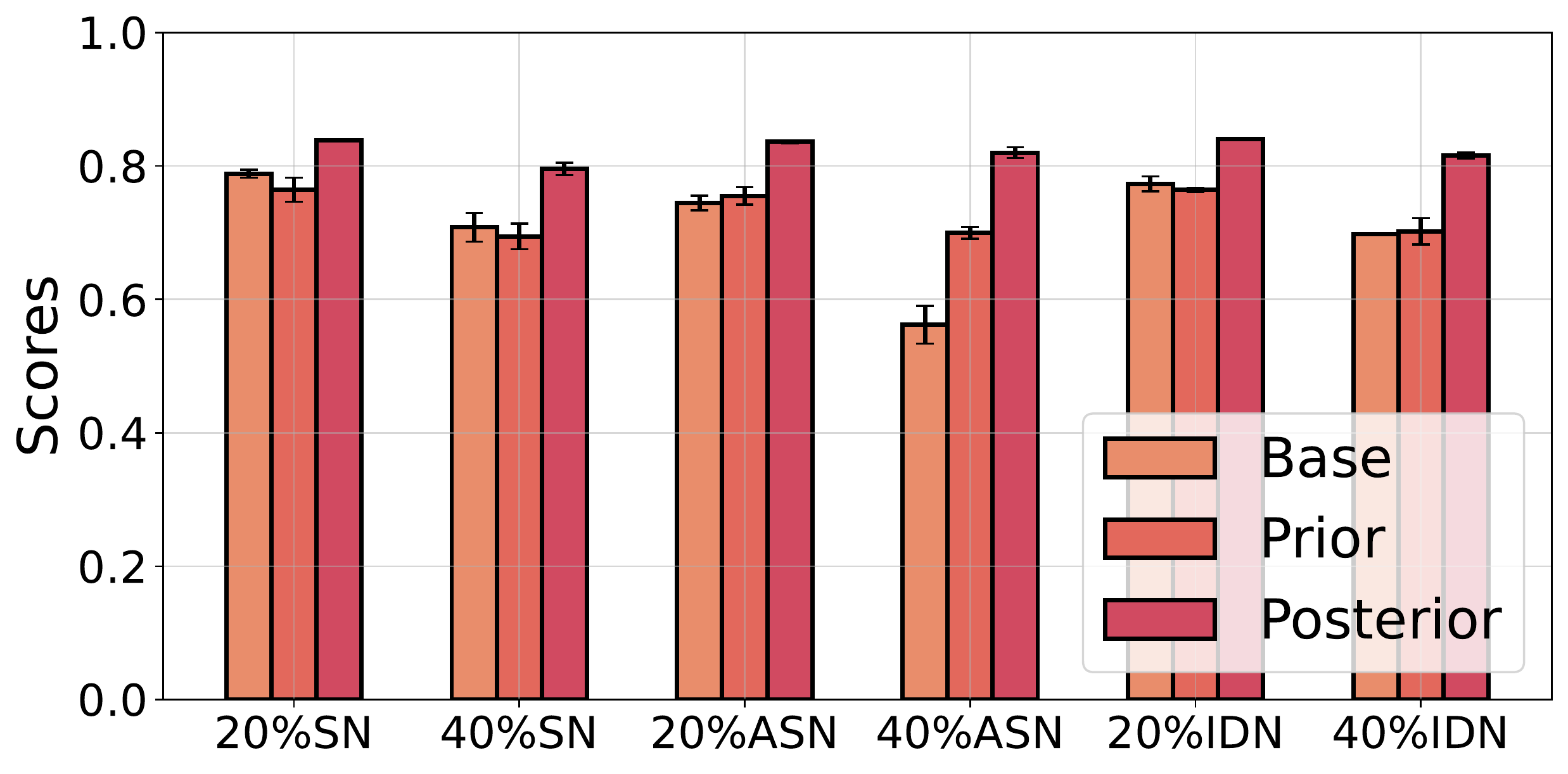}
    \vspace{-2ex}
  \caption{Performance comparison on 20newsgroup dataset between applying prior or posterior for true label prediction.}
  \vspace{-2ex}
  \label{fig:posterior}
\end{figure}

\begin{figure*}[!ht]
	\centering
	\vspace{-2ex}
	\subfigure[SN]{
		\includegraphics[width=0.31\linewidth]{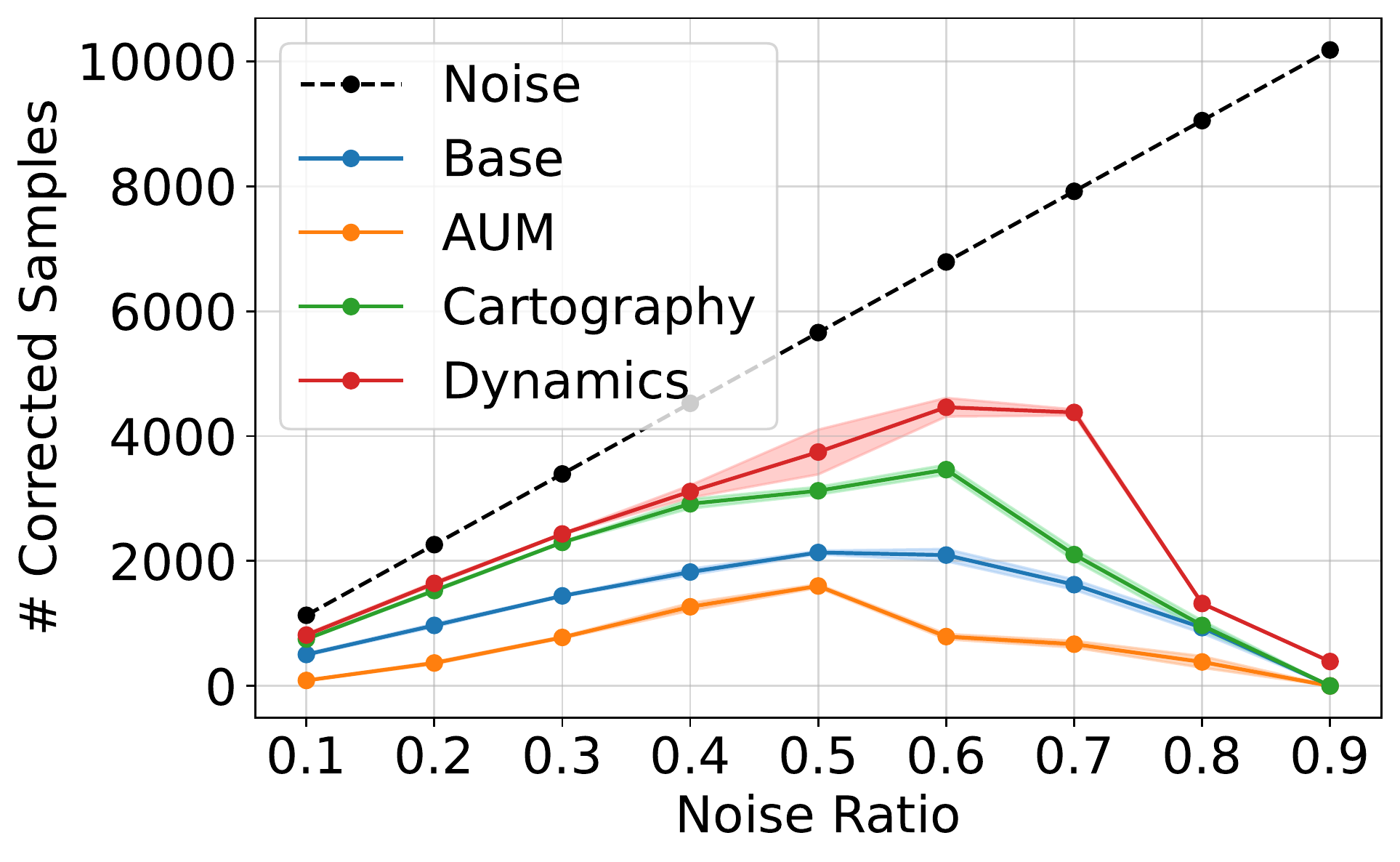}
		\label{fig:pq-sn}
	} 
	\subfigure[ASN]{
		\includegraphics[width=0.31\linewidth]{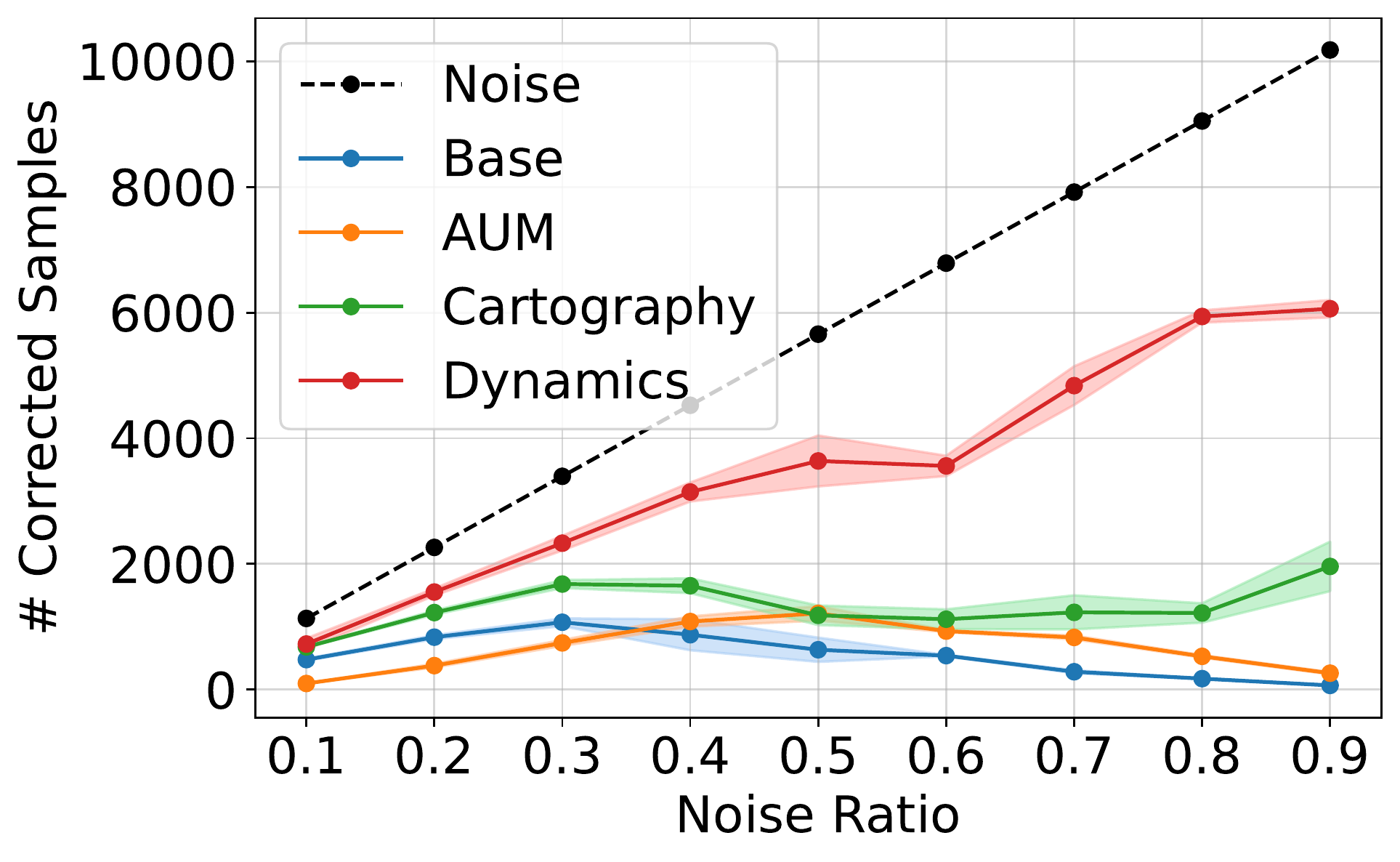}
		\label{fig:pq-asn}
	}  
	\subfigure[IDN]{
		\includegraphics[width=0.31\linewidth]{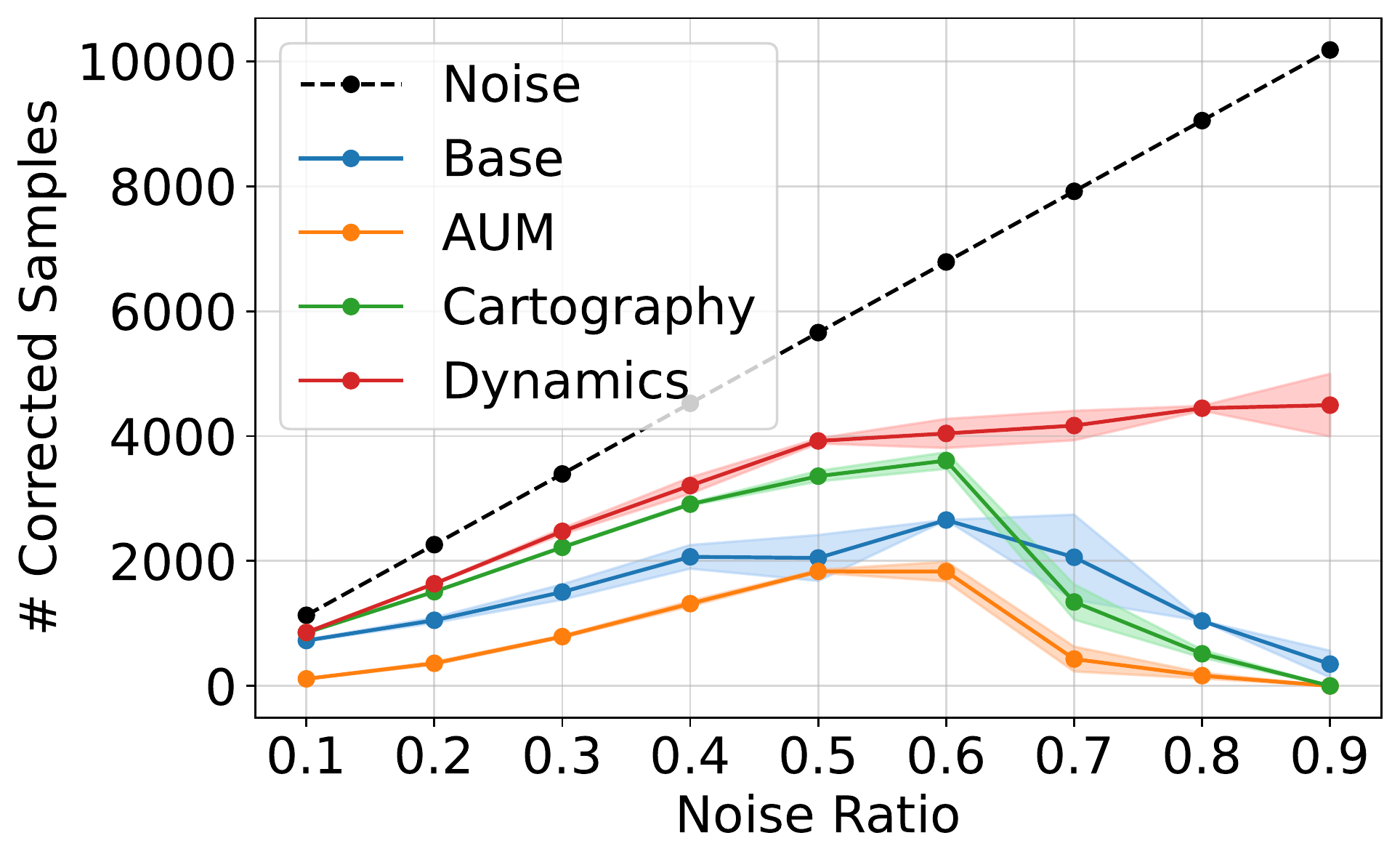}
		\label{fig:pq-idn}
	}  
	\vspace{-3ex}
	\caption{Numbers of samples corrected by various training dynamic patterns as prior on 20newsgroup dataset.}\label{fig:prior-qual}
\end{figure*}

\begin{figure*}[t]
	\centering
	\vspace{-2ex}
	\subfigure[SN]{
		\includegraphics[width=0.31\linewidth]{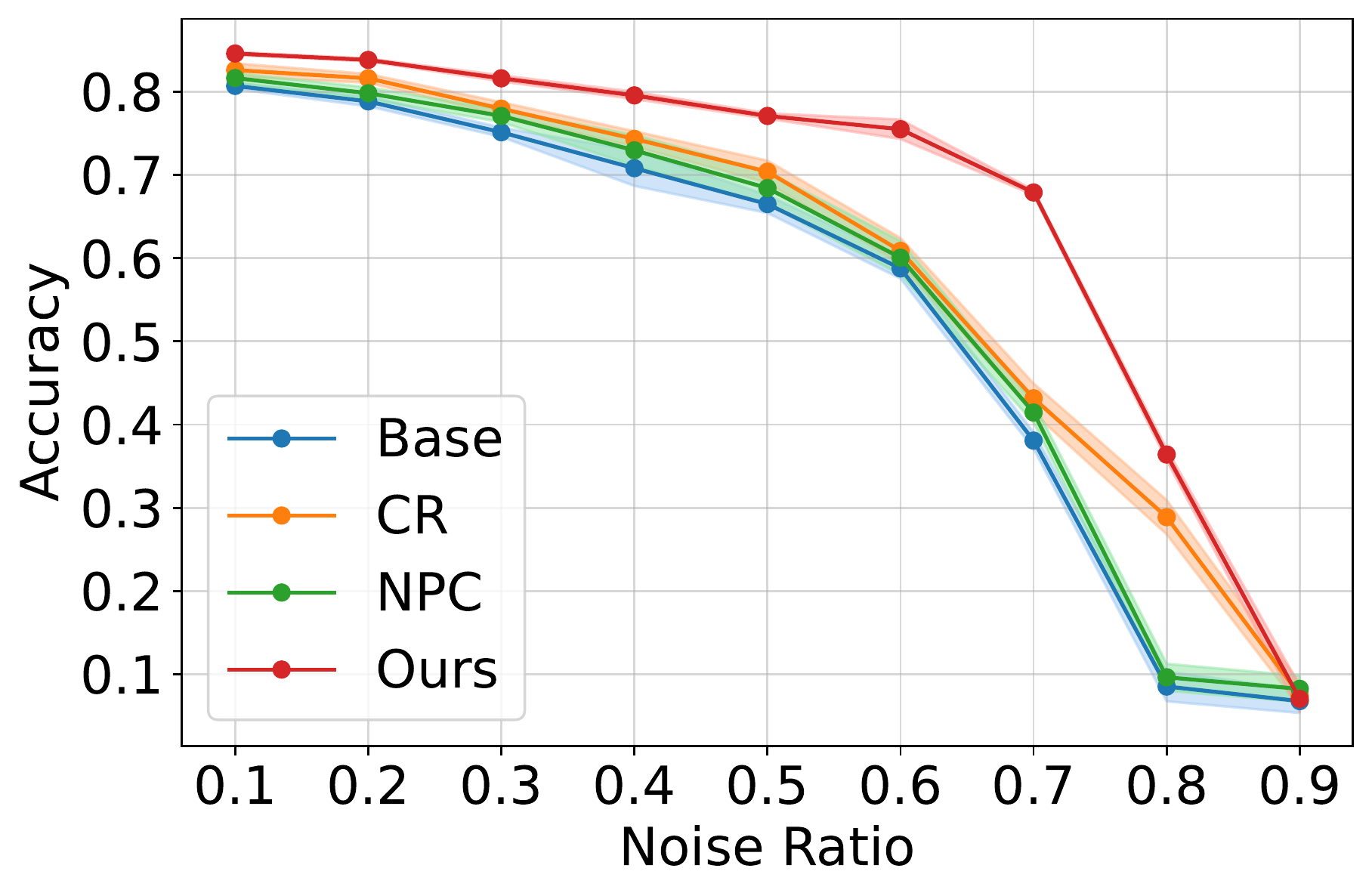}
		\label{fig:le-sn}
	}  
	\subfigure[ASN]{
		\includegraphics[width=0.31\linewidth]{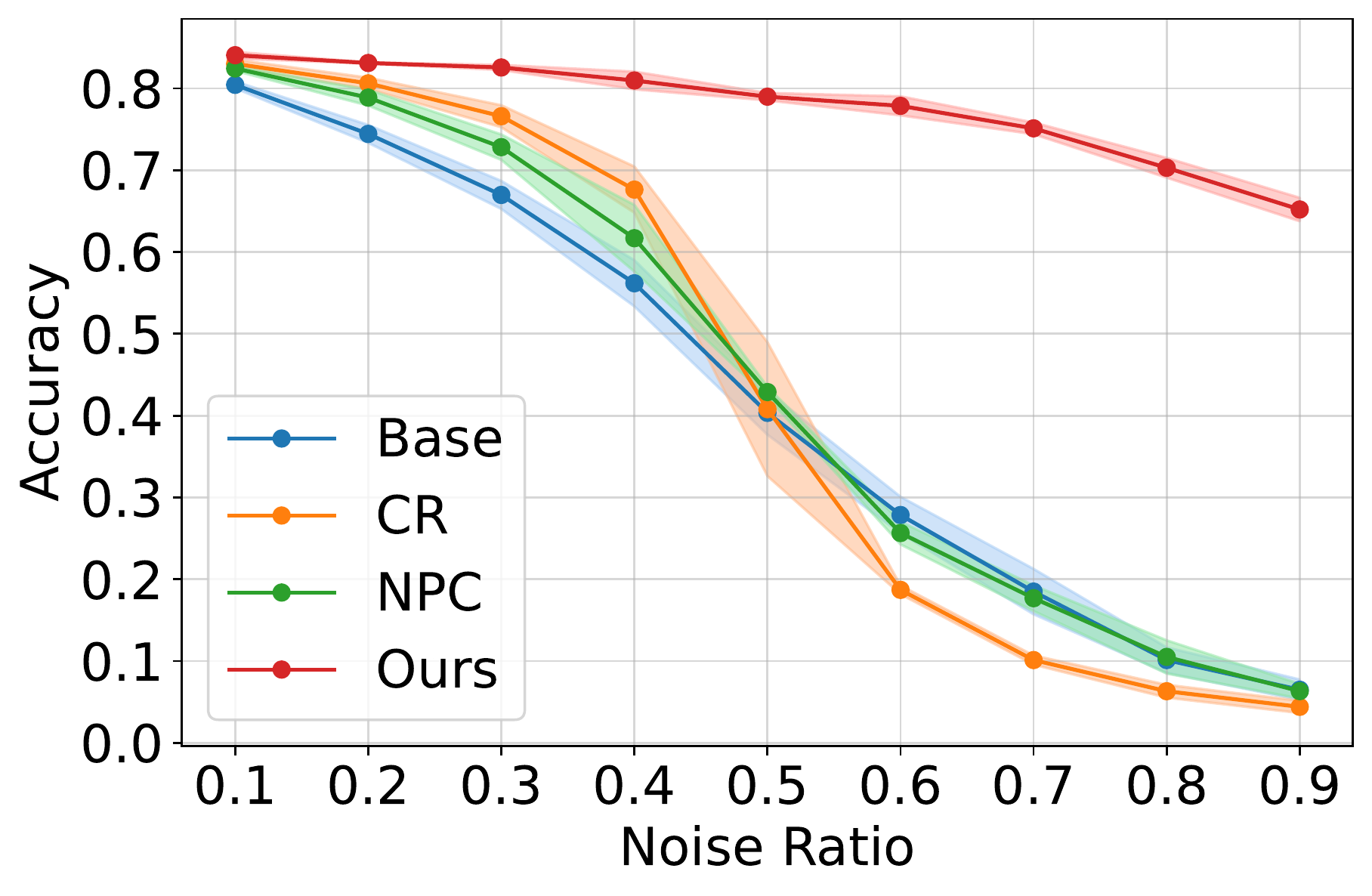}
		\label{fig:le-asn}
	}  
	\subfigure[IDN]{
		\includegraphics[width=0.31\linewidth]{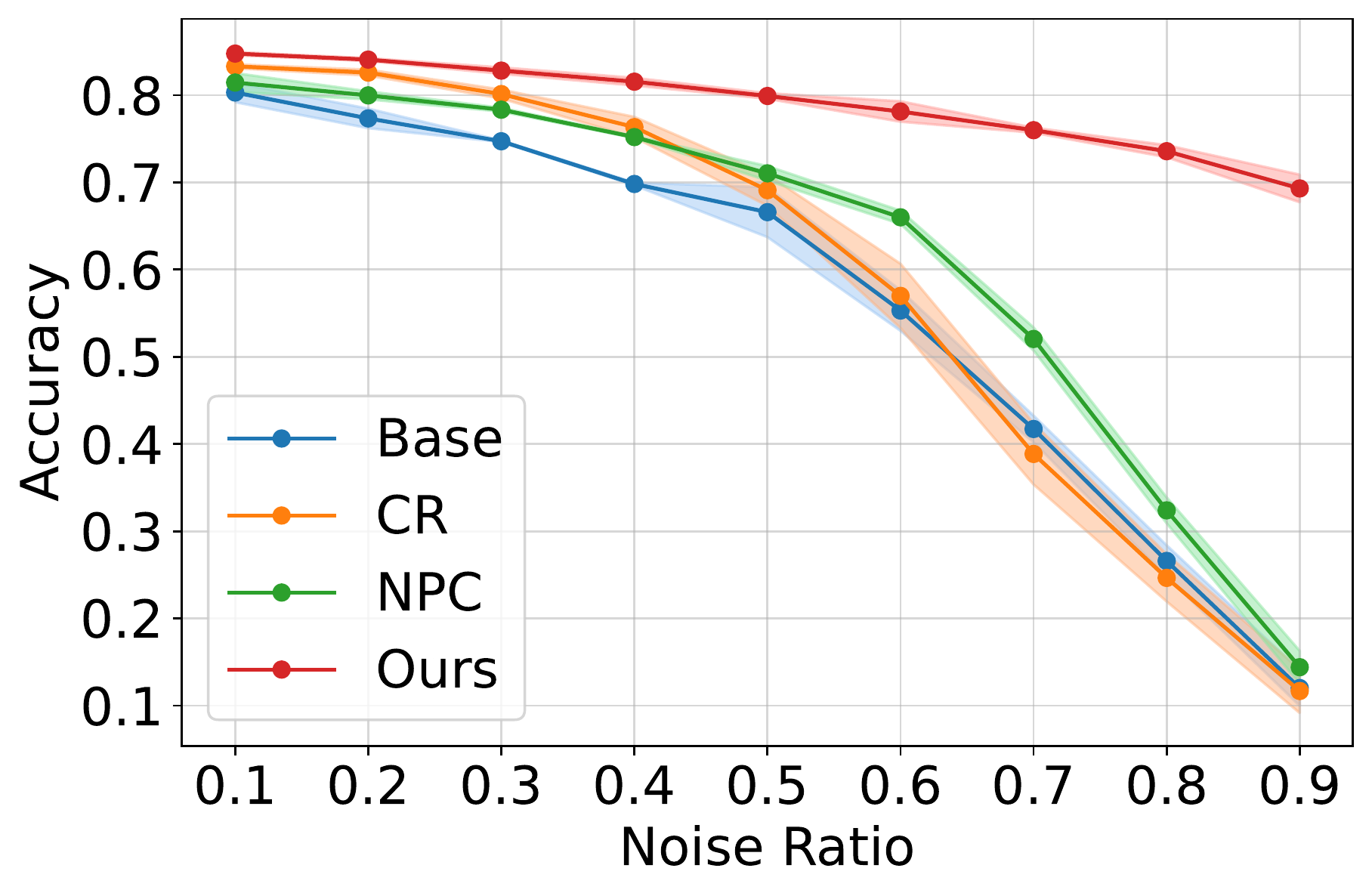}
		\label{fig:le-idn}
	}  
	\vspace{-2ex}
	\caption{Performance curves under different noise ratios on 20newsgroup dataset.}\label{fig:large}
	\vspace{-1ex}
\end{figure*}

\input{sections/53ablation.tex}


\input{sections/52prior.tex}

\input{sections/54large.tex}
\input{sections/54quality.tex}
\input{sections/56case.tex}

%% file: sections/50setup.tex
\subsection{Experimental Setup}\label{subsec:exp-setup}
\subsubsection{Datasets}
To verify the efficacy of \ours, we first experiment on two synthetic-noise datasets: 20newsgroup~\cite{LANG1995331} and AG NEWS~\cite{li-roth-2002-learning, zhang2015character}.
Three different types of synthetic label noise are generated and injected into the datasets following the setups of existing works on learning from noisy supervision~\cite{Patrini_2017_CVPR, yao2020dual, NEURIPS2021_23451391, pmlr-v162-bae22a}:
(1) \textbf{Symmetric Noise (SN)} flips labels uniformly to other classes~\cite{Patrini_2017_CVPR, NEURIPS2018_a19744e2, xia2021robust, pmlr-v162-bae22a};
(2) \textbf{Asymmetric Noise (ASN)} flips labels within similar classes~\cite{Tanaka_2018_CVPR, NEURIPS2018_a19744e2, xia2021robust, pmlr-v162-bae22a};
and (3) \textbf{Instance-Dependent Noise (IDN)} flips label with a probability proportional to the features of the sample~\cite{xia2020part, cheng2021learning, NEURIPS2021_23451391, pmlr-v162-bae22a}.
Furthermore, we conduct experiments on three real-world datasets: ChemProt~\cite{Krallinger2017OverviewOT}, TREC~\cite{Awasthi2020Learning}, and SemEval~\cite{zhou2020nero}.
ChemProt is a chemical-protein interaction dataset with $10$ classes;
TREC is a question classification dataset with $6$ classes;
and SemEval is a relation extraction dataset with $9$ classes.
For these three datasets, we use the  pre-defined heuristic rules from prior works~\cite{yu-etal-2021-fine,zhang2wrench,zhang2022prompt} as weak supervision to generate noisy labels. 
The noise ratio of ChemProt, TREC, and SemEval are $22.88\%$, $38.56\%$, and $16.00\%$, respectively.
See Appendix.~\ref{app:dataset} for details.

\subsubsection{Baselines}
We compare with the most relevant state-of-the-art noisy label learning baselines from different categories: 
(1) \textit{Basic Performances} without specific design for robust learning with label noise~\cite{devlin-etal-2019-bert};
(2) \textit{Multi-Model Training Strategies}, including \textbf{Co-Teaching}~\cite{NEURIPS2018_a19744e2}, \textbf{JoCoR}~\cite{Wei_2020_CVPR}, \textbf{CR}~\cite{zhou-chen-2021-learning};
(3) \textit{Generative Models for Noisy Matrix Estimation}, including \textbf{DualT}~\cite{yao2020dual}, \textbf{CausalNL}~\cite{NEURIPS2021_23451391}, \textbf{NPC}~\cite{pmlr-v162-bae22a}, and \textbf{CR w/ NPC}.
See Appendix.~\ref{app:baseline} for details.

\subsubsection{Evaluation Protocol}
All experiments are evaluated using accuracy on a clean test set, and the reported test performance is selected according to the performance on a clean development set.
This applies to both \ours and all baselines.
We report the average performance as well as standard deviations using 5 random seeds.

\subsubsection{Implementation Details}
We implement \ours using PyTorch~\cite{paszke2019pytorch} and HuggingFace~\cite{wolf-etal-2020-transformers}.
In the experiments on ChemProt, BioBERT~\cite{lee2020biobert} is used as the backbone model for the first stage training dynamics pattern encoder $\theta$, while for the rest datasets, we use BERT~\cite{devlin-etal-2019-bert}.
We use the same backbone for all baseline methods.
See Appendix.~\ref{app:imp} for more details.

%% file: tables/tab-syn.tex
\begin{table*}[!t]
\caption{Main results on synthetic noise datasets.}
\label{tab:main_syn}
\vspace{-2ex}
\centering 
\renewcommand\arraystretch{0.88}
\fontsize{7.5}{9.5}\selectfont \setlength{\tabcolsep}{0.3em}
\resizebox{0.99\linewidth}{!}{%
\begin{tabular}{l|cccccc|cccccc}
\toprule
\bf    Dataset ($\rightarrow$)     & \multicolumn{6}{c|}{\textbf{20newsgroup}}  &  \multicolumn{6}{c}{\textbf{AG NEWS}}  \\  \hline   
\bf    Method ($\downarrow$) / Noise ($\rightarrow$)    & \bf  $20\%$ SN & \bf $40\%$ SN & \bf $20\%$ ASN  & \bf $40\%$ ASN & \bf $20\%$ IDN & \bf $40\%$ IDN & \bf  $20\%$ SN & \bf $40\%$ SN & \bf $20\%$ ASN  & \bf $40\%$ ASN & \bf $20\%$ IDN & \bf $40\%$ IDN\\ \hline
\multicolumn{10}{l}{\emph{Basic Performances}}  \\\hline 
Base & 78.84\std{0.59} & 70.81\std{2.13} & 74.44\std{1.09} &56.18\std{2.82}	& 77.33\std{1.12} & 69.81\std{0.16} & 82.08\std{1.80} & 78.17\std{2.69} & 81.43\std{0.92} & 77.15\std{2.50} & 85.59\std{1.91}& 75.86\std{3.09}	\\ \hline
\multicolumn{10}{l}{\emph{Multi-Model Training Strategies}}  \\\hline 
Co-Teaching & 77.66\std{1.06} & 69.25\std{3.62} & 77.51\std{1.84} & 67.26\std{1.94} &  77.45\std{0.32} & 73.76\std{1.63} & 82.99\std{1.15}& 78.79\std{2.41} & 81.96\std{0.77} & 78.07\std{1.68} & 87.85\std{0.82}& 76.52\std{2.21}	 \\ 
JoCoR & 80.92\std{0.64} & 73.27\std{4.25} & 81.01\std{0.20} & 69.40\std{3.39} & 81.57\std{0.76} & 74.19\std{0.83} & 83.82\std{1.08}& 81.26\std{1.83} & 85.88\std{0.98} & 77.98\std{1.04} & 87.04\std{1.27} & 79.93\std{1.46}	  \\
CR & 81.61\std{0.53} & 74.33\std{0.92} &80.62\std{0.69} & 67.63\std{2.82} & 82.58\std{0.34} & 76.33\std{1.17} & 89.10\std{1.86}& 78.40\std{0.4}& 89.03\std{1.20}& 74.52\std{1.29} & 87.48\std{2.17} & 75.29\std{1.11}	 \\
\hline
\multicolumn{10}{l}{\emph{Generative Models for Noisy Transition Matrix Estimation}} 
 \\\hline
DualT & 78.92\std{0.56} & 73.39\std{1.19} & 74.66\std{1.24} & 67.82\std{2.33} & 77.16\std{0.76} & 70.61\std{0.56} & 83.66\std{0.97} & 80.84\std{2.05} & 82.11\std{1.27}& 79.03\std{2.59} & 86.47\std{1.55}&	78.84\std{2.45}\\
CausalNL & 81.08\std{0.54} & 74.43\std{1.93} & 81.22\std{1.04} & 71.25\std{2.70} & 82.57\std{0.64} & 78.91\std{2.33}	& 86.44\std{1.98}&82.74\std{2.11} &89.87\std{1.64} & 79.80\std{2.22} & 89.00\std{1.67} & 84.62\std{2.58}	\\
NPC & 79.82\std{0.70} & 72.96\std{1.84} &78.88\std{0.97} &61.69\std{4.07} & 79.97\std{0.46} & 75.19\std{0.11} & 82.83\std{3.43} & 75.04\std{5.53} & 83.94\std{1.97}& 77.69\std{2.87}& 86.28\std{1.17}&	77.38\std{4.09}
\\
CR w/ NPC & 83.09\std{0.11} & 77.96\std{1.00} &83.13\std{0.39} & 73.50\std{3.61} & 83.47\std{0.16} & 80.47\std{0.71} & 89.69\std{0.53} & 83.21\std{1.06} & 89.01\std{1.22} & 82.54\std{2.69} & 88.25\std{1.49}& 86.41\std{1.93}	\\
\rowcolor[HTML]{EFEFEF} \bf {\ours}  (Our Method) & \bf{83.82\std{0.04}} & \bf 79.56\std{0.93}& \bf  83.63\std{0.23} & \bf  81.98\std{0.80} & \bf 84.07\std{0.17}  & \bf  81.54\std{0.44} & \bf 91.42\std{0.70}& \bf 89.80\std{0.58} & \bf 91.37\std{0.98} & \bf 90.43\std{1.11} & \bf 91.41\std{0.49} & \bf 88.90\std{1.66}	\\
\bottomrule
\end{tabular}
}
\vspace{-2ex}
\end{table*}

%% file: tables/tab-real.tex
\begin{table}[tb]
\centering
\caption{Main results on real-world noise datasets.}\label{tab:main_real}
\vspace{-2ex}
\fontsize{7.5}{9.5}\selectfont \setlength{\tabcolsep}{0.5em}
\begin{tabular}{@{}lccc@{}}
\toprule
\bf Method            & \bf ChemProt & \bf TREC & \bf SEMEVAL\\ \midrule
Noise Ratio & 22.88$\%$          &  38.56$\%$       &16.00$\%$ \\\midrule
Base        & 64.84\std{0.28}    &  67.33\std{0.83} & 71.44\std{0.10}       \\
Co-Teaching & 65.98\std{0.63}    &  66.61\std{0.35}  & 72.07\std{0.76}      \\
JoCoR       & 65.32\std{0.24}    &  66.50\std{1.44}  & 70.33\std{1.10}      \\
CR          & 65.53\std{0.22}    &  68.33\std{0.31}  & 71.11\std{1.07}      \\
DualT       & 65.30\std{2.18}    &  69.33\std{1.02}  & 70.88\std{1.07}     \\
CausalNL    & 67.29\std{1.37}    &  69.83\std{2.71}  & 72.22\std{0.26}       \\
NPC         & 65.15\std{0.51}    &  70.44\std{0.39}  & 72.17\std{0.17}     \\
CR w/ NPC   & 66.46\std{0.23}    &  71.02\std{0.43}  & 72.72\std{0.70}       \\
\rowcolor[HTML]{EFEFEF} \bf \ours  (Our Method)          & \bf 69.07\std{0.38}    &  \bf  72.39\std{0.82} & \bf 73.17\std{0.29}\\ \bottomrule
\end{tabular}
\vspace{-1ex}
\end{table}

%% file: sections/51main.tex
\vspace{-0.5ex}
\subsection{Main Results}

\noindent \textbf{Performance Comparison.} Table~\ref{tab:main_syn} and Table~\ref{tab:main_real} show the main results for the synthetic and the real-world noisy datasets.
From these results, we have the following observations:

\noindent (1) \ours significantly outperforms all the baselines on synthetic datasets with varying noise types and ratios.
Additionally, \ours also shows superiority in performance on real-world noisy datasets.
Compared with the strongest baseline, CR w/ NPC, directly concatenating the Co-Regularized classifier with the generative model for noisy prediction calibration, \ours achieves $3.10\%$ gain on average on synthetic datasets and $1.48\%$ on real-world datasets.

\noindent (2) The results show that \ours has larger performance gains on synthetic datasets compared to real-world datasets.
This is because real-world datasets often contain a more intricate mix of noises, making it a bit of a challenge for the model to arrive at accurate estimates.
Additionally, the noise ratio on real-world noisy datasets is lower than that of synthetic datasets, resulting in less room for improvement with our proposed method.

\noindent (3) Compared to the 20newsgroup dataset, the gains of Noisy Transition Matrix estimation-based methods over other baselines are more pronounced for the AG News dataset. 
This discrepancy can be attributed to the difference in the number of classes between the two datasets. 
Specifically, the AG News dataset has only four classes, which is much smaller than 20newsgroup. This makes the estimation of the corresponding transition matrix simpler. 
Our observation suggests that the ability to better estimate the transition matrix leads to improved performance in the AG News dataset.


\noindent \textbf{Model Calibration.}
As a probabilistic denoising method, we find that \ours also improves model calibration when fine-tuning PLMs with noisy labels~\cite{kong2020calibrated}.
Figure~\ref{fig:calibration} shows 
the calibration results of the strong CR baseline and \ours on the 20newsgroup dataset corrupted with $40\%$ symmetric noise.
The results suggest that while CR shows some robustness in noisy label learning, it suffers from under-confidence.
This is because some model branches with lower confidences will regularize the other ones, leading to an also lower average for prediction.
In contrast, \ours not only improves the classification accuracy on the clean test set but also better calibrates the predictions and 
reduces the expected calibration error (ECE) from $27.12\%$ to $13.93\%$ (see details in Appendix.~\ref{app:calibration}).


%% file: sections/53ablation.tex
\subsection{Ablation Study}

\noindent \textbf{Effect of Different Components.} To investigate the effectiveness of each model component, 
we remove some components of the model to test how the performance varies:
(1) Removing the co-regularization mechanism from the Stage I model, degenerating to a simple noisy-supervised classifier;
(2) Removing the training dynamics from the dynamics-based prior function, leading to the degeneration to a vanilla KNN prior function;
(3) Removing the co-regularization mechanism from the Stage II model, resulting in the degeneration to the original NPC architecture.
Table~\ref{tab:ablation} shows the impact of removing components from \ours on both synthetic (20newsgroup) and real-world (ChemProt) datasets.
The results reveal that as more components are taken away, the performance of the model deteriorates, emphasizing the significant contribution of each component to the overall performance.
The co-regularization mechanism in the second stage proves to be more effective when it is also applied in the first stage.
This is because multiple branches of the co-regularized second stage model generate consistent $q_\phi(y|\hat{y},\mathbf{x})$ estimates based on the same input $p(\hat{y}|\mathbf{x})$ and prior knowledge.

\input{tables/tab-ablation.tex}

\noindent \textbf{Comparing Prior and Posterior.}
Figure~\ref{fig:posterior} compares the performance of the posterior distribution of the generative model against that of the prior distribution used directly for posterior inference.
It indicates that while using prior for inference can produce comparable or even better results than the noisy-supervised classifier in the first stage, it still falls short compared to the posterior produced by the generative model in the second stage.
This highlights the importance of the second stage, which uses a co-regularized generative model for calibrating noisy predictions and refining the imperfect prior knowledge to only retain the key information.

%% file: tables/tab-ablation.tex
\begin{table}[t]
\vspace{-0.3ex}
\centering 
\caption{Ablation studies in \ours on synthetic noise dataset 20newsgroup. ``\textbf{I}'' indicates the stage I model; ``\textbf{P}'' represents the prior function; ``\textbf{II}'' refers to the stage II model.}
\vspace{-1ex}
\label{tab:ablation}
\renewcommand\arraystretch{0.88}
\fontsize{7.5}{9.5}\selectfont \setlength{\tabcolsep}{0.2em}
\resizebox{0.99\linewidth}{!}{%
\begin{tabular}{ccc|ccccccc}
\toprule
\bf   I & \bf P & \bf II   & \bf  $20\%$ SN & \bf $40\%$ SN & \bf $20\%$ ASN  & \bf $40\%$ ASN & \bf $20\%$ IDN & \bf $40\%$ IDN &\bf ChemProt\\ \hline
\XSolidBrush & \XSolidBrush & \XSolidBrush & 78.84\std{0.59} & 70.81\std{2.13} & 74.44\std{1.09} &56.18\std{2.82}	& 77.33\std{1.12} & 69.81\std{0.16} & 65.15\std{0.51}\\\hline
\XSolidBrush & \XSolidBrush & \Checkmark & 79.77\std{0.51} & 72.93\std{0.20} & 78.84\std{1.01} & 61.76\std{4.19} &  80.00\std{0.37} & 75.20\std{0.08} & 67.05\std{0.40}\\ 
\XSolidBrush & \Checkmark & \XSolidBrush & 81.44\std{0.31} & 77.06\std{0.04} & 82.05\std{0.66} & 79.49\std{0.49} & 81.34\std{0.22} & 77.99\std{0.44} & 65.03\std{0.51}\\
\Checkmark & \XSolidBrush & \XSolidBrush & 83.09\std{0.11} & 77.96\std{1.00} & 83.13\std{0.39} & 73.50\std{3.61} & 83.47\std{0.16} & 80.68\std{0.62} &65.46\std{0.22}\\ \hline
\XSolidBrush & \Checkmark & \Checkmark & 81.64\std{0.06} & 76.34\std{0.92} & 82.05\std{0.66} & 79.51\std{0.57} & 81.37\std{0.23} & 78.00\std{0.65} &67.23\std{0.84}\\
\Checkmark & \Checkmark & \XSolidBrush & 83.58\std{0.16} & 79.67\std{0.21} & 83.04\std{0.5} & 81.54\std{0.82} & 83.35\std{0.62} & 81.08\std{0.49} &67.44\std{0.31}\\
\Checkmark & \XSolidBrush & \Checkmark & 83.56\std{0.10} & 78.45\std{1.05} & 83.57\std{0.57} & 75.22\std{0.04} & 83.46\std{0.39} & 80.89\std{0.35}	& 66.48\std{0.34}
\\\hline
\Checkmark & \Checkmark & \Checkmark & \bf 83.82\std{0.04} & \bf 79.90\std{0.51} & \bf 83.63\std{0.23} & \bf 82.31\std{0.23} & \bf 84.07\std{0.17} & \bf 81.54\std{0.44}&\bf 69.07\std{0.38}\\
\bottomrule
\end{tabular}
}
\vspace{-1ex}
\vspace{-2ex}
\end{table}

%% file: sections/52prior.tex
\subsection{Quality Analysis of Prior}

To evaluate the quality of the dynamics-based prior (\textbf{Dynamics}) in \ours, we compare with a set of state-of-the-art training dynamic patterns and treat them as the prior knowledge:
(1) \textbf{Base}~\cite{pmlr-v162-bae22a} is the original KNN-based prior used in NPC;
(2) \textbf{AUM}~\cite{NEURIPS2020_c6102b37} measures the average difference between the logit values for a sample's assigned class and its highest non-assigned class;
(3) \textbf{Cartography}~\cite{swayamdipta-etal-2020-dataset} observes that instances with smaller mean and standard deviations on output probabilities during the training process often correspond to labeling errors.
Figure~\ref{fig:prior-qual} shows the numbers of samples corrected with different prior knowledge. 
The results demonstrate that our proposed dynamics-based prior consistently achieves superior performance across varying noise types and ratios, highlighting its effectiveness in supplying high-quality true-label information for the subsequent generative model for noisy prediction calibration.

%% file: sections/54large.tex
\subsection{Performance with Large Noise Ratio}

Figure~\ref{fig:large} displays the evaluation of models under large noise ratios ($\geq 50\%$).
The results demonstrate the robustness of \ours to large noise ratios.
Besides, we can also observe that \ours shows increased performance gains, compared with the other methods, as the magnitude of label noise increases.
This observation also exists in Table~\ref{tab:main_syn}.
As is also shown in Figure~\ref{fig:prior-qual}, this can be attributed to its dynamics-based prior function, which provides high-quality true-label information to the generative model in the second stage.

%% file: sections/6conclusion.tex
\section{Conclusion}
In this paper, we focus on leveraging training dynamics to correct noisy predictions by considering the larger distances between noisy samples and their assigned label clusters.
In comparison to clean samples, the noisy samples consistently exhibit a higher mean and deviation in distances throughout 1,500 experiments, resulting in a noticeable discrepancy in their training patterns during PLM fine-tuning.
To enhance the quality of prior knowledge and improve robustness to noisy labels, we propose \ours, a framework for noisy label learning that integrates training dynamics patterns with deep generative models.
We leverage the agreement from multiple branches optimized by the co-regularization loss, as opposed to solely relying on potentially unreliable noisy labels. 
Our proposed method, \ours, demonstrates an average accuracy improvement of $2.55\%$ on five benchmark datasets with both synthetic and real-world noise.
Moreover, we conducted extensive experiments to validate the effectiveness of each component.
We believe that this study opens up new possibilities in the topics of using training trajectories to handle noisy labels, especially in calibrating noisy predictions under large noise ratios.

%% file: appendices/a-datasets.tex
\section{Dataset Details}\label{app:dataset}
\begin{table}[!h]
\centering
\renewcommand\arraystretch{0.88}
\fontsize{7.5}{9.5}\selectfont \setlength{\tabcolsep}{0.3em}
\caption{Detailed dataset statistics.}\label{tab:dataset}
\begin{tabular}{@{}lcccc@{}}
\toprule
Datasets    & \# Training & \# Validation & \# Test & \# Label \\ \midrule
20newsgroup & 9051        & 2263          & 7532    & 20       \\
AG NEWS     & 40000       & 7600          & 7600    & 4        \\
ChemProt    & 12861       & 1607          & 1607    & 10       \\
TREC        & 4965        & 500           & 500     & 6        \\
SemEval     & 1749        & 200           & 692     & 9        \\
WOS         & 22552       & 5639          & 18794   & 134      \\ \bottomrule
\end{tabular}
\end{table}

In this work, we select \textbf{20newgroup}~\cite{LANG1995331} and \textbf{AG NEWS}~\cite{li-roth-2002-learning, zhang2015character} for news topic classification for experiments on synthetic noise datasets.
We also conducted experiments on an additional synthetic noise dataset, \textbf{WOS}~\cite{kowsari2017HDLTex}. The results of these experiments are available in ~\cref{app:WOS}.
Table.~\ref{tab:dataset} introduces detailed statistics about datasets used in our experiments with both synthetic and real-world noise.
Since these datasets are assumed to have no noisy labels, we manage three types of noisy label for noisy label injection.
In this part, we explain the details of these noisy generation processes.
Since we include a detailed explanation of symmetric and asymmetric noise in \cref{subsec:exp-setup}, we will emphasize on the instance-dependent noise (IDN) in the following.
Specifically, we follow the noise generation process in existing literature~\cite{xia2020part,cheng2021learning, NEURIPS2021_23451391, pmlr-v162-bae22a} for IDN generation in our experiments. The detailed algorithm of IDN noisy label generation is summarized in Alg.~\ref{alg:idn}:

\begin{algorithm}[!ht]
	\begin{small}
	\KwIn{Clean samples $(\mathbf{x}_i,y_i)_{i=1}^n$; Noise ratio $\tau$.}
   Sample instance flip rates $q\in\mathbb{R}^{n}$ from the truncated normal distribution $\mathcal{N}(\tau,0.1^2,[0,1]).$\\
   Independently sample $\mathbf{v}_1,\mathbf{v}_2,\cdots,\mathbf{v}_c$ from the standard normal distribution $\mathcal{N}(0,1^2)$.\\
	\For{$e = 1, 2, \cdots, n$}{ 
		{
		 $p=\mathbf{x}_i\times \mathbf{v}_{y_i}$.\\
		  $p_{y_i}=-\inf$.\\
            $p=q_i\times{\rm softmax}(p)$. \\
            $p_{y_i}=1-q_i$. \\
            Randomly choose a label from the label space according to probabilities $p$ as noisy label $\tilde{y}_i$.
		}
	}
	\KwOut{Noisy samples $(\mathbf{x}_i,\tilde{y}_i)_{i=1}^n$.}
	\end{small}
	\caption{Instance Dependent Noise Generation. }
	\label{alg:idn}
\end{algorithm}

In addition, we conduct experiments on three real-world datasets, including: \textbf{ChemProt}~\cite{Krallinger2017OverviewOT} for chemical-protein interaction classification, \textbf{TREC}~\cite{Awasthi2020Learning} for question classification, and \textbf{SemEval}~\cite{zhou2020nero} for relation extraction. The generation of noisy labels on TREC and ChemProt datasets are introduced in Appendix.~\ref{app:rule}. We follow the same process as \cite{zhou2020nero} to generate noisy labels  for SemEval.

\section{Additional Results on WOS}
\label{app:WOS}

Table.~\ref{tab:main_wos} displays the additional experimental result on \textbf{WOS}~\cite{kowsari2017HDLTex}.

\begin{table}[!ht]
\caption{Main results on synthetic noise dataset WOS.}
\label{tab:main_wos}
\vspace{-0.3ex}
\centering 
\renewcommand\arraystretch{0.88}
\fontsize{7.5}{9.5}\selectfont \setlength{\tabcolsep}{0.3em}
\resizebox{0.99\linewidth}{!}{%
\begin{tabular}{l|cccccc}
\toprule
\bf    Dataset ($\rightarrow$) &  \multicolumn{6}{c}{\textbf{WOS}}  \\  \hline   
\bf    Method ($\downarrow$)    & \bf  $20\%$ SN & \bf $40\%$ SN & \bf $20\%$ ASN  & \bf $40\%$ ASN & \bf $20\%$ IDN & \bf $40\%$ IDN \\ \hline
\multicolumn{7}{l}{\emph{Basic Performances}}  \\\hline 
Base & 78.35\std{0.14} & 76.01\std{0.82} & 78.03\std{0.24}& 65.13\std{0.42}& 78.19\std{0.21} & 76.36\std{0.31}	\\ \hline
\multicolumn{7}{l}{\emph{Multi-Model Training Strategies}}  \\\hline 
Co-Training & 79.31\std{0.53} & 77.79\std{1.60}&79.10\std{0.87} & 67.61\std{1.45} & 78.43\std{0.39}&	76.03\std{1.35} \\ 
JoCoR & 78.45\std{0.63}& 77.96\std{0.42} & 78.46\std{0.41}& 66.38\std{0.92} & 78.49\std{0.49} &	 76.47\std{1.36}\\
CR & 78.25\std{0.13} & 76.35\std{0.17}& 77.41\std{0.82}& 62.32\std{4.38}&77.48\std{0.93} & 75.47\std{2.03}	 \\
\hline
\multicolumn{7}{l}{\emph{Generative Models for Noisy Transition Matrix Estimation}} 
 \\\hline
DualT & 78.99\std{0.98}& 76.74\std{1.22}& 78.58\std{0.68}& 67.27\std{0.79}& 78.35\std{0.29}&76.79\std{0.71}	\\
CausalNL & 78.56\std{1.06}& 76.97\std{0.88} & 79.19\std{1.25}& 65.43\std{0.75}& 78.46\std{0.59} & 76.53\std{0.96}	\\
NPC & 78.80\std{0.34} & 77.80\std{0.86}& 79.21\std{0.80}& 68.36\std{0.30}& 78.86\std{0.25}& 77.50\std{0.51}
\\
CR w/ NPC & 79.18\std{0.47} & 78.15\std{0.98}&79.28\std{0.33} & 70.79\std{0.96} & 79.34\std{0.42} & 77.87\std{1.13}	\\
\rowcolor[HTML]{EFEFEF} \bf {\ours}& \bf 79.95\std{0.10} & \bf  78.68\std{0.26} & \bf 79.55\std{0.23} & \bf 72.50\std{3.17}& \bf 79.62\std{0.19}& \bf 78.09\std{0.37}	\\
\bottomrule
\end{tabular}
}
\end{table}

%% file: appendices/a-rules.tex
\section{Details for Weak Supervision}\label{app:rule}
\begin{table*}[!ht]\small
    \centering
    \caption{Examples of semantic rules on Chemprot.}
    \label{tab:chem}
    \begin{tabular}
    {p{210pt} | p{260pt}}
    \toprule
         \textbf{Rule} & \textbf{Example}\\
         \midrule
         \texttt{HAS (x, [amino acid,mutant, mutat, replace] ) $\rightarrow$ part\_of } 
         & A major part of this processing requires endoproteolytic cleavage at specific pairs of basic [CHEMICAL]amino acid[CHEMICAL] residues, an event necessary for the maturation of a variety of important biologically active proteins, such as insulin and [GENE]nerve growth factor[GENE].\\ \hline
         
         \texttt{HAS (x, [bind, interact, affinit] ) $\rightarrow$ regulator} &
         The interaction of [CHEMICAL]naloxone estrone azine[CHEMICAL] (N-EH) with various [GENE]opioid receptor[GENE] types was studied in vitro.\\ \hline
         
         \texttt{HAS (x, [activat, increas, induc, stimulat, upregulat] ) $\rightarrow$ upregulator/activator }
         & The results of this study suggest that [CHEMICAL]noradrenaline[CHEMICAL] predominantly, but not exclusively, mediates contraction of rat aorta through the activation of an [GENE]alphalD-adrenoceptor[GENE]. \\ \hline
        
        \texttt{HAS (x, [downregulat, inhibit, reduc, decreas] ) $\rightarrow$ downregulator/inhibitor }
        & These results suggest that [CHEMICAL]prostacyclin[CHEMICAL] may play a role in downregulating [GENE]tissue factor[GENE] expression in monocytes, at least in part via elevation of intracellular levels of cyclic AMP.\\ \hline
        
        \texttt{HAS (x, [ agoni, tagoni]* ) $\rightarrow$ agonist}  * (note the leading whitespace in both cases)
         & Alprenolol and BAAM also caused surmountable antagonism of [CHEMICAL]isoprenaline[CHEMICAL] responses, and this [GENE]beta 1-adrenoceptor[GENE] antagonism was slowly reversible. \\ \hline
         
         \texttt{HAS (x, [antagon] ) $\rightarrow$ antagonist }
         & It is concluded that [CHEMICAL]labetalol[CHEMICAL] and dilevalol are [GENE]beta 1-adrenoceptor[GENE] selective antagonists. \\ \hline
         
         \texttt{HAS (x, [modulat, allosteric] ) $\rightarrow$ modulator}
         & [CHEMICAL]Hydrogen sulfide[CHEMICAL] as an allosteric modulator of [GENE]ATP-sensitive potassium channels[GENE] in colonic inflammation. \\ \hline
         
         \texttt{HAS (x, [cofactor] ) $\rightarrow$ cofactor}
         & The activation appears to be due to an increase of [GENE]GAD[GENE] affinity for its cofactor, [CHEMICAL]pyridoxal phosphate[CHEMICAL] (PLP). \\ \hline
         
         \texttt{HAS (x, [substrate, catalyz, transport, produc, conver] ) $\rightarrow$ substrate/product}
         & Kinetic constants of the mutant [GENE]CrAT[GENE] showed modification in favor of longer [CHEMICAL]acyl-CoAs[CHEMICAL] as substrates. \\ \hline
         
         \texttt{HAS (x, [not] ) $\rightarrow$ not}
          & [CHEMICAL]Nicotine[CHEMICAL] does not account for the CSE stimulation of [GENE]VEGF[GENE] in HFL-1. \\
    \bottomrule
    \end{tabular}
\end{table*}

The examples of semantic rules on ChemProt are given in Table.~\ref{tab:chem}. 

%% file: appendices/a-baseline.tex
\section{Baselines Details} \label{app:baseline}
We compare with the most relevant state-of-the-art baselines on learning with noisy labels, including: 
(1) \textbf{Base}~\cite{devlin-etal-2019-bert} is the BERT$_{\rm base}$ model fine-tuned only with standard cross-entropy loss and early stopping;
(2) \textbf{Co-Teaching}~\cite{NEURIPS2018_a19744e2} trains two different networks and feeds the samples with small loss to each other for parameter updating;
(3) \textbf{JoCoR}~\cite{Wei_2020_CVPR} also trains two networks and selects the samples, for which the sum of the losses from two networks is small, as clean samples;
(4) \textbf{CR}~\cite{zhou-chen-2021-learning} is another method of training multiple networks and uses a soft target to regularize each model;
(5) \textbf{DualT}~\cite{yao2020dual} factorizes the transition probability matrix into the product of two independently-estimated matrices to mitigate the error in the estimation;
(6) \textbf{CausalNL}~\cite{NEURIPS2021_23451391} introduces an auxiliary latent variable $\mathbf{z}$, generating $\mathbf{x}$ together with $y$, and proposes a structural causal model for instance-dependent noise learning;
(7) \textbf{NPC}~\cite{pmlr-v162-bae22a} proposes a deep generative model to estimate the transition from noisy predictions to true labels;
(8) \textbf{CR w/ NPC} treats NPC as a post-processing module and links it after the CR method, which achieves the best post-performance of NPC-based methods we obtain empirically.

%% file: appendices/a-implementation.tex
\section{Implementation Details}\label{app:imp}

\begin{table}[!ht]
\centering
\fontsize{7.5}{9.5}\selectfont \setlength{\tabcolsep}{0.3em}
\caption{Hyper-parameter configurations.}\label{tab:hyper}
\begin{tabular}{@{}lcccccc@{}}
\toprule
                     & 20newsgroup & AG NEWS & WOS  & ChemProt & TREC & SemEval \\ \midrule
Max Length           & 150         & 256     & 256  & 512      & 64   & 128     \\
Batch Size           & 64          & 32      & 32   & 16       & 64   & 32      \\
Learning Rate        & 1e-4        & 1e-4    & 1e-4 & 2e-5     & 5e-6 & 5e-6    \\
$\#$ Stage I Epochs    & \multicolumn{6}{c}{10}\\
$\#$ Stage II Iterations &  \multicolumn{6}{c}{10}\\
($\lambda_1$, $\lambda_2$, $\delta$, $\rho$)      & \multicolumn{6}{c}{(5,1,1,2)} \\\bottomrule
\end{tabular}
\end{table}

All experiments are conducted on CPU: Intel(R) Core(TM) i7-5930K CPU @ 3.50GHz and GPU: NVIDIA GeForce RTX A5000 GPUs using python 3.8 and PyTorch 1.12. 
Table.~\ref{tab:hyper} shows the detailed hyper-parameter configuration.
In addition, we search the batch size in $\{8,16,32,64\}$ and use Adam~\cite{kingma2014adam} as optimizer.

%% file: appendices/a-calibration.tex
\section{Model Calibration}\label{app:calibration}
Machine learning applications usually require trustworthy predictions that need to be not only accurate but also well-calibrated.
Specifically, a well-calibrated model is expected to output prediction confidence comparable to its classification accuracy.
More precisely, for a data point $\mathbf{x}_i$, we denote by $y_i$ the ground-truth label, $\hat{y}_i$  the prediction made by the model, and $\hat{p}(\mathbf{x})$ the output probability associated with the prediction.
The calibration error of the predictive model for a given confidence $p\in(0,1)$ is defined as:
\begin{equation}
\begin{aligned}
\epsilon_p=|\mathbb{P}(\hat{y}=y|\hat{p}(\mathbf{x})=p)-p|.
\end{aligned}\label{eq:calibration1}
\end{equation}
We partition all data points into $M$ bins of equal size according to their prediction confidences bounded $l_m$ and $u_m$.
Then for any $p\in[l_m,u_m)$, we define the empirical calibration error~\cite{pmlr-v70-guo17a} as:
\begin{equation}
\begin{aligned}
\hat{\epsilon}_p=\hat{\epsilon}_m=\frac{1}{|\mathcal{B}_m|}\sum_{i\in\mathcal{B}_m}[\mathbbm{1}(\hat{y}_i=y_i)-\hat{p}_i]|,
\end{aligned}
\end{equation}
where $y_i$, $\hat{y}_i$, and $\hat{p}_i$ are the true label, prediction, and confidence of sample $i$.
$|\mathcal{B}_m|$ is the sample size of $m$-th bin.
To evaluate the overall calibration error of the predictive model, we can further take a weighted average of the calibration errors of all bins, which is also known as the \textbf{Expected calibration error (ECE)}~\cite{naeini2015obtaining}.
The metric is defined as:
\begin{equation}
\begin{aligned}
\text{ECE}=\sum_{m=1}^M\frac{|B_m|}{N}|\text{acc}(B_m)-\text{conf}(B_m)|=\sum_{m=1}^M\frac{|B_m|}{N}\hat{\epsilon}_m,
\end{aligned}
\end{equation}
where $N$ is the number of samples. Specifically, we set $M=10$.

%% file: sections/58parameter.tex
\subsection{Parameter Study}
\begin{figure}[t]
  \centering
  \includegraphics[width=0.9\linewidth]{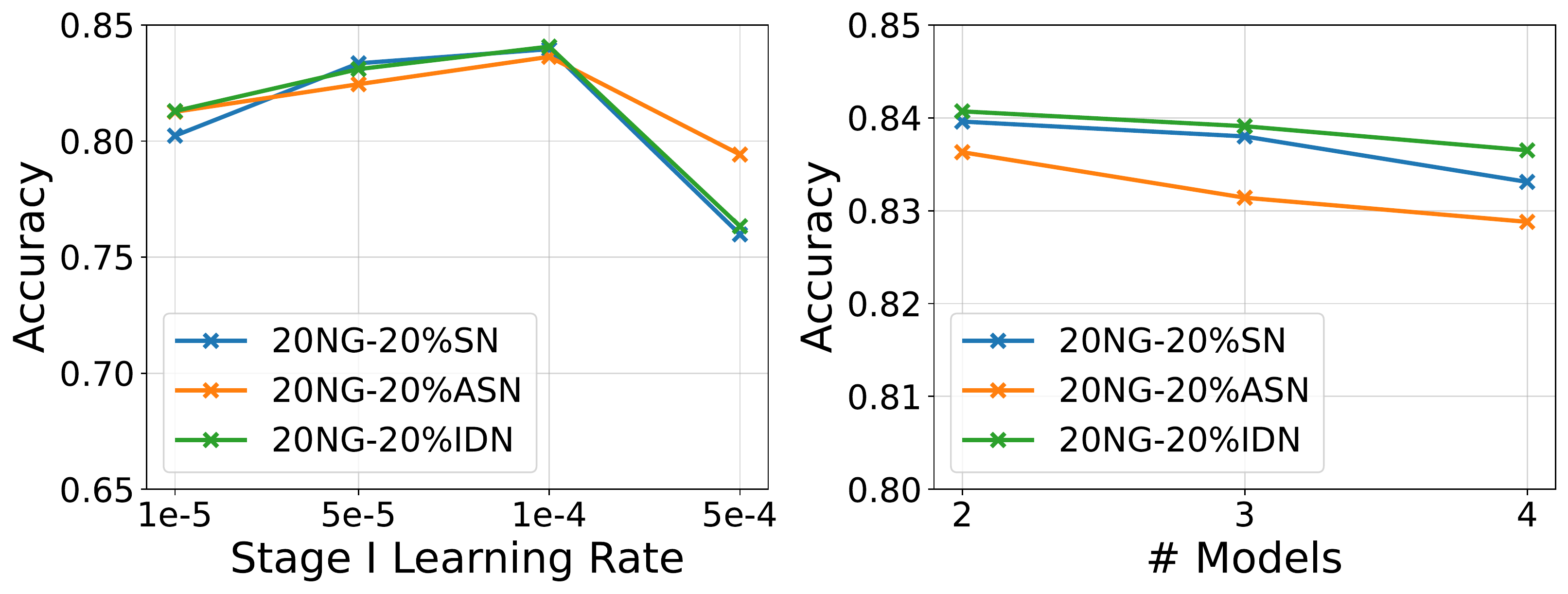}
  \vspace{-3ex}
  \caption{Parameter studies on learning rate of training dynamic pattern encoding and the number of model branches under different noise types on 20newsgroup dataset.}
  \label{fig:parameter}
  \vspace{-3ex}
\end{figure}

We conduct experiments to investigate the impact of two hyper-parameters on the performance of \ours: the learning rate of the training pattern encoding, which may affect the quality of the prior, and the number of branches in the co-regularization mechanism.
The other hyper-parameters remain the same as the default. 

\noindent \textbf{Effect of Stage I Learning Rate.}
The choice of the learning rate plays a crucial role in encoding the training dynamics patterns.
When the learning rate is set too low, the training trajectory pattern may be extended during the learning process, leading to a relatively slight impact on the model performance. On the other hand, if the learning rate is set too high, the pattern may change too rapidly, causing overfitting to the noisy labels and a decrease in performance.
Furthermore, a high learning rate can also produce low-quality probability distributions $p(\hat{y}|\mathbf{x})$, hindering the ability of the subsequent model to perform noisy prediction calibration.

\noindent\textbf{Effect of Model Branches.}
Generally, the number of branches does not severely influence the model performance.
We hypothesize that including more branches could be gradually more difficult for models to reach a consensus on incoming noisy samples. 